\documentclass{article}

\usepackage{microtype}
\usepackage{graphicx}
\usepackage{subfigure}
\usepackage{booktabs} 

\usepackage{hyperref}

\usepackage[accepted]{icml}

\usepackage{amsmath}
\usepackage{amssymb}
\usepackage{mathtools}
\usepackage{amsthm}
\usepackage{bm}

\usepackage[capitalize,noabbrev]{cleveref}

\theoremstyle{plain}
\newtheorem{theorem}{Theorem}[section]
\newtheorem{proposition}[theorem]{Proposition}
\newtheorem{lemma}[theorem]{Lemma}

\theoremstyle{definition}

\theoremstyle{remark}
\newtheorem{remark}[theorem]{Remark}

\usepackage[textsize=tiny]{todonotes}

\icmltitlerunning{Online Pseudo-Zeroth-Order Training of Neuromorphic Spiking Neural Networks}

\begin{document}

\twocolumn[
\icmltitle{Online Pseudo-Zeroth-Order Training of Neuromorphic Spiking Neural Networks}

\icmlsetsymbol{equal}{*}

\begin{icmlauthorlist}
\icmlauthor{Mingqing Xiao}{pku}
\icmlauthor{Qingyan Meng}{cuhk,sz}
\icmlauthor{Zongpeng Zhang}{pkub}
\icmlauthor{Di He}{pku,pkuai}
\icmlauthor{Zhouchen Lin}{pku,pkuai,pengcheng}
\end{icmlauthorlist}

\icmlaffiliation{pku}{National Key Lab of General AI, School of Intelligence Science and Technology, Peking University}
\icmlaffiliation{cuhk}{The Chinese University of Hong Kong, Shenzhen}
\icmlaffiliation{sz}{Shenzhen Research Institute of Big Data}
\icmlaffiliation{pkub}{Department of Biostatistics, School of Public Health, Peking University}
\icmlaffiliation{pkuai}{Institute for Artificial Intelligence, Peking University}
\icmlaffiliation{pengcheng}{Peng Cheng Laboratory}

\icmlcorrespondingauthor{Mingqing Xiao}{mingqing\_xiao@pku.edu.cn}
\icmlcorrespondingauthor{Zhouchen Lin}{zlin@pku.edu.cn}

\icmlkeywords{Neuromorphic Computing, Spiking Neural Networks, Non-Backpropagation Training, Pseudo-Zeroth-Order}

\vskip 0.3in
]



\printAffiliationsAndNotice{}  

\begin{abstract}
Brain-inspired neuromorphic computing with spiking neural networks (SNNs) is a promising energy-efficient computational approach. However, successfully training SNNs in a more biologically plausible and neuromorphic-hardware-friendly way is still challenging. Most recent methods leverage spatial and temporal backpropagation (BP), not adhering to neuromorphic properties. Despite the efforts of some online training methods, tackling spatial credit assignments by alternatives with comparable performance as spatial BP remains a significant problem. In this work, we propose a novel method, online pseudo-zeroth-order (OPZO) training. Our method only requires a single forward propagation with noise injection and direct top-down signals for spatial credit assignment, avoiding spatial BP's problem of symmetric weights and separate phases for layer-by-layer forward-backward propagation. OPZO solves the large variance problem of zeroth-order methods by the pseudo-zeroth-order formulation and momentum feedback connections, while having more guarantees than random feedback. Combining online training, OPZO can pave paths to on-chip SNN training. Experiments on neuromorphic and static datasets with fully connected and convolutional networks demonstrate the effectiveness of OPZO with similar performance compared with spatial BP, as well as estimated low training costs.
\end{abstract}

\vspace{-4mm}
\section{Introduction}
\label{sec:intro}

Neuromorphic computing with biologically inspired spiking neural networks (SNNs) is an energy-efficient computational framework with increasing attention recently~\citep{roy2019towards,schuman2022opportunities}. Imitating biological neurons to transmit spike trains for sparse event-driven computation as well as parallel in-memory computation, efficient neuromorphic hardware is developed, supporting SNNs with low energy consumption~\citep{davies2018loihi,pei2019towards,wozniak2020deep,rao2022long}.

Nevertheless, supervised training of SNNs is challenging considering neuromorphic properties. While popular surrogate gradient methods can deal with the non-differentiable problem of discrete spikes~\citep{shrestha2018slayer,wu2018spatio,neftci2019surrogate}, they rely on backpropagation (BP) through time and across layers for temporal and spatial credit assignment, which is biologically problematic and would be inefficient on hardware.

Particularly, spatial BP suffers from problems of weight transport and separate forward-backward stages with update locking~\citep{crick1989recent,frenkel2021learning}, and temporal BP is further infeasible for spiking neurons with the online property~\citep{bellec2020solution}. Considering learning in biological systems with unidirectional local synapses, maintaining reciprocal forward-backward connections with symmetric weights and separate phases of signal propagation is often viewed as biologically problematic~\citep{nokland2016direct}, and also poses challenges for efficient on-chip training of SNNs. Methods with only forward passes, or with direct top-down feedback signals acting as modulation in biological three-factor rules~\citep{fremaux2016neuromodulated,roelfsema2018control}, are more efficient and plausible, e.g., on neuromorphic hardware~\citep{davies2021taking}.

Some previous works explore alternatives for temporal and spatial credit assignment. To deal with temporal BP, online training methods are developed for SNNs~\citep{bellec2020solution,xiao2022online}. With tracked eligibility traces, they decouple temporal dependency and support forward-in-time learning. However, alternatives to spatial BP still require deeper investigations. Most existing works mainly rely on random feedback~\citep{nokland2016direct,bellec2020solution}, but have limited guarantees and poorer performance than spatial BP. Some works explore forward gradients~\citep{silver2022learning,baydin2022gradients}, but they require an additional stage of heterogeneous signal propagation and usually perform poorly due to the large variance. Recently, \citet{malladi2023fine} show that zeroth-order (ZO) optimization with simultaneous perturbation stochastic approximation (SPSA) can effectively fine-tune pre-trained large language models, but the method requires specially designed settings, as well as two forward passes, and does not work for general neural network training due to the large variance. On the other hand, local learning has been studied, e.g., with local readout layers~\citep{kaiser2020synaptic} or forward-forward self-supervised learning~\citep{hinton2022forward,ororbia2023learning}. It is complementary to global learning and can improve some methods~\citep{ren2023scaling}. As a crucial component of machine learning, efficient global learning alternatives with competitive performance remain an important problem.

In this work, we propose a novel online pseudo-zeroth-order (OPZO) training method with only a single forward propagation and direct top-down feedback for global learning. We first propose a pseudo-zeroth-order formulation for neural network training, which decouples the model function and the loss function, and maintains the zeroth-order formulation for neural networks while leveraging the available first-order property of the loss function for more informative feedback error signals. Then we propose momentum feedback connections to directly propagate feedback signals to hidden layers. The connections are updated based on the one-point zeroth-order estimation of the expectation of the Jacobian, with which the large variance of zeroth-order methods can be solved and more guarantees are maintained compared with random feedback. OPZO only requires a noise injection in the common forward propagation, flexibly applicable to black-box or non-differentiable models. Built upon online training, OPZO enables training in a similar form as the three-factor Hebbian learning based on direct top-down modulations, paving paths to on-chip training of SNNs. 
Our contributions include: 
\begin{enumerate}
    \item We propose a pseudo-zeroth-order formulation that decouples the model and loss function for neural network training, which enables more informative feedback signals while keeping the zeroth-order formulation of the (black-box) model. 
    \item We propose the OPZO training method with a single forward propagation and momentum feedback connections, solving the large variance of zeroth-order methods and keeping low costs. Built on online training, OPZO provides a more biologically plausible method friendly for potential on-chip training of SNNs.
    \item We conduct extensive experiments on neuromorphic and static datasets with both fully connected and convolutional networks, as well as on ImageNet with larger networks finetuned under noise. Results show the effectiveness of OPZO in reaching similar or superior performance compared with spatial BP and its robustness under different noise injections. OPZO is also estimated to have lower computational costs than BP on potential neuromorphic hardware.
\end{enumerate}

\section{Related Work}\label{sec:related_work}

\paragraph{SNN Training Methods}
A mainstream method to train SNNs is spatial and temporal BP combined with surrogate gradient (SG)~\citep{shrestha2018slayer,wu2018spatio,neftci2019surrogate,li2021differentiable} or gradients with respect to spiking times~\citep{zhang2020temporal,kim2020unifying,zhu2022training}. Another direction is to derive equivalent closed-form transformations or implicit equilibriums between specific encodings of spike trains, e.g., (weighted) firing rates or the first time to spike, and convert artificial neural networks (ANNs) to SNNs~\citep{rueckauer2017conversion,deng2021optimal,stockl2021optimized,meng2022trainingnn} or directly train SNNs with gradients from the equivalent transformations~\citep{lee2016training,zhou2021temporal,wu2021tandem,meng2022training} or equilibriums~\citep{o2019training,xiao2021training,martin2021eqspike,xiao2022spide}. To tackle the problem of temporal BP, some online training methods are proposed~\citep{bellec2020solution,xiao2022online,bohnstingl2022online,meng2023towards,yin2023accurate} for forward-in-time learning, but most of them still require spatial BP. Considering alternatives to spatial BP, \citet{neftci2017event,lee2020spike,bellec2020solution} apply random feedback, \citet{kaiser2020synaptic} propose online local learning, and \citet{yang2022training} propose local tandem learning with ANN teachers. Different from them, we propose a new method for global learning while maintaining similar performance as spatial BP and much better results than random feedback.

\citet{li2021differentiable} and \citet{mukhoty2023direct} study zeroth-order properties for each parameter or neuron to adjust surrogate functions or leverage a local zeroth-order estimator for the Heaviside step function, lying in the spatial and temporal BP framework. Differently, in this work, zeroth-order training refers to simultaneous perturbation for global network training without spatial BP.

\vspace{-3mm}
\paragraph{Alternatives to Spatial Backpropagation}
For effective and more biologically plausible global learning of neural networks, some alternatives to spatial BP are proposed. Target propagation~\citep{lee2015difference}, feedback alignment (FA)~\citep{lillicrap2016random}, and sign symmetric~\citep{liao2016important,xiao2018biologically} avoid the weight symmetric problem by propagating targets or using random or only sign-shared backward weights, and \citet{akrout2019deep} improves random weights by learning it to be symmetric with forward weights. They, however, still need an additional stage of sequential layer-by-layer backward propagation. Direct feedback alignment (DFA)~\citep{nokland2016direct,launay2020direct} improves FA to directly propagate errors from the last layer to hidden ones. However, random feedback methods have limited guarantees and perform much worse than BP. Some recent works study forward gradients~\citep{silver2022learning,baydin2022gradients,ren2023scaling,bacho2024low}, but they require an additional heterogeneous signal propagation stage for forward gradients and may suffer from biological plausibility issues and larger costs. 
There are also methods to train neural networks with energy functions~\citep{scellier2017equilibrium} or use lifted proximal formulation~\citep{li2020training}. Besides global supervision, some works turn to local learning, using local readout layers~\citep{kaiser2020synaptic} or forward-forward contrastive learning~\citep{hinton2022forward}. This work mainly focuses on global learning and can be combined with local learning.

\vspace{-2mm}
\paragraph{Zeroth-Order Optimization} ZO optimization has been widely studied in machine learning, but its application to direct neural network training is limited due to the variance caused by a large number of parameters. ZO methods have been used for black-box optimization~\citep{grill2015black}, adversarial attacks~\citep{chen2017zoo}, reinforcement learning~\citep{salimans2017evolution}, etc., at relatively small scales. For neural network training with extremely high dimensions, recently, \citet{yue2023zeroth} theoretically show that the complexity of ZO optimization can exhibit weak dependencies on dimensionality considering the effective dimension, and \citet{malladi2023fine} propose zeroth-order SPSA for memory-efficient fine-tuning pre-trained large language models with a similar theoretical basis. However, it requires specially designed settings (e.g., fine-tuning under the prompt setting is important for the effectiveness~\citep{malladi2023fine}) which is not applicable to general neural network training, as well as two forward passes. \citet{jiang2023one} propose a likelihood ratio method to train neural networks, but it requires multiple forward propagation proportional to the layer number in practice. \citet{chen2023deepzero} consider the finite difference for each parameter rather than simultaneous perturbation and propose pruning methods for improvement, limited in computational complexity. Differently, this work proposes a pseudo-zeroth-order method for neural network training from scratch with only one forward pass in practice for low costs and comparable performance to spatial BP.

\section{Preliminaries}

\subsection{Spiking Neural Networks}\label{sec:snn}

Imitating biological neurons, each spiking neuron keeps a membrane potential $u$, integrates input spike trains, and generates a spike for information transmission once $u$ exceeds a threshold. $u$ is reset to the resting potential after a spike. We consider the commonly used leaky integrate and fire (LIF) model with the dynamics of the membrane potential as: $\tau_m\frac{du}{dt} = -(u-u_{rest}) + R\cdot I(t), \text{for}\, u < V_{th}$, with input current $I$, threshold $V_{th}$, resistance $R$, and time constant $\tau_m$. 
When $u$ reaches $V_{th}$ at time $t^f$, the neuron generates a spike and resets $u$ to zero. The output spike train is $s(t) = \sum_{t^f}\delta(t-t^f)$. 

SNNs consist of connected spiking neurons. We consider the simple current model $I_i(t)=\sum_j w_{ij}s_j(t) +b_i$, where $i,j$ represent the neuron index, $w_{ij}$ is the weight and $b_i$ is a bias. The discrete computational form is: 
\vspace{-2mm}
\begin{equation}
    \left\{
    \begin{aligned}
        &u_i\left[t + 1\right] = \lambda (u_i[t] - V_{th}s_i[t]) + \sum_j w_{ij}s_j[t] + b_i,\\
        &s_i[t + 1] = H(u_i\left[t+1\right] - V_{th}).\\
    \end{aligned}
    \right.
    \label{eq.discrete}
\end{equation}
Here $H(x)$ is the Heaviside step function,  $s_i[t]$ is the spike signal at discrete time step $t$, and $\lambda<1$ is a leaky term (taken as $1-\frac{1}{\tau_m}$). 
For multi-layer networks, we use $\mathbf{s}^{l+1}[t]$ to represent the $(l+1)$-th layer’s response after receiving signals $\mathbf{s}^l[t]$ from the $l$-th layer, i.e., the expression is $\mathbf{u}^{l+1}[t+1]=\lambda(\mathbf{u}^{l+1}[t]-V_{th}\mathbf{s}^{l+1}[t])+\mathbf{W}^l\mathbf{s}^l[t+1]+\mathbf{b}^l$.

\paragraph{Online Training of SNNs} 
We build the proposed OPZO on online training methods for forward-in-time learning. Here online training refers to online through the time dimension of SNNs~\citep{bellec2020solution,xiao2022online}, as opposed to backpropagation through time. We consider OTTT~\citep{xiao2022online} to online calculate gradients at each time by the tracked presynaptic trace $\hat{\mathbf{a}}^l[t] = \sum_{\tau \leq t}\lambda^{t-\tau}\mathbf{s}^l[\tau]$ and instantaneous gradient $\mathbf{g}_{\mathbf{u}^{l+1}}[t]=\left(\frac{\partial \mathcal{L}[t]}{\partial \mathbf{s}^N[t]}\prod_{i=0}^{N-l-2}\frac{\partial \mathbf{s}^{N-i}[t]}{\partial \mathbf{s}^{N-i-1}[t]} \frac{\partial \mathbf{s}^{l+1}[t]}{\partial \mathbf{u}^{l+1}[t]}\right)^\top$ as $\nabla_{\mathbf{W}^l}\mathcal{L}[t]=\mathbf{g}_{\mathbf{u}^{l+1}}[t]{\hat{\mathbf{a}}^l[t]}^\top$. In OTTT, the instantaneous gradient requires layer-by-layer spatial BP with surrogate derivatives for ${\frac{\partial \mathbf{s}^{l}[t]}{\partial \mathbf{u}^{l}[t]}}$. The proposed OPZO, on the other hand, leverages only one forward propagation across layers and direct feedback to estimate $\mathbf{g}_{\mathbf{u}^{l+1}}[t]$ without spatial BP combining surrogate gradients.

\begin{figure*} [ht]
    \centering
    \includegraphics[width=\textwidth]{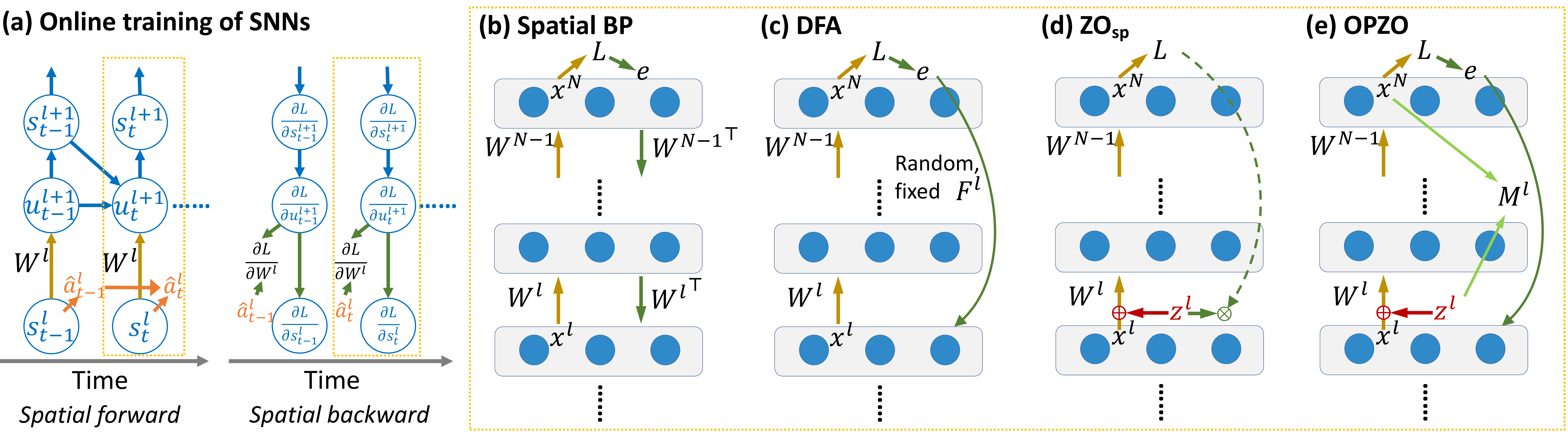}
    \vspace{-6mm}
    \caption{Illustration of different training methods. (a) Online training of SNNs with tracked traces for temporal credit assignment~\citep{bellec2020solution,xiao2022online}. (b-e) Different spatial credit assignment methods. (b) Spatial BP with SG propagates errors layer-by-layer with symmetric weights. (c) DFA~\citep{nokland2016direct} directly propagates error signals from the top layer to the middle ones with fixed random connections. (d) Single-point zeroth-order methods add perturbation during forward propagation, and afterward, the loss signal is passed to the middle layers. (e) The proposed OPZO method leverages momentum feedback connections based on perturbation vectors and directly propagates error signals to neurons with top-down connections.}
    \label{fig:illustration}
    \vspace{-3mm}
\end{figure*}

\subsection{Zeroth-Order Optimization}\label{sec:zo}

Zeroth-order optimization is a gradient-free method using only function values. A classical ZO gradient estimator is SPSA~\citep{spall1992multivariate}, which estimates the gradient of parameters $\bm{\theta}$ for $\mathcal{L}(\bm{\theta})$ on a random direction $\mathbf{z}$ as:
\begin{equation}
    \nabla^{ZO}\mathcal{L}(\bm{\theta}) = \frac{\mathcal{L}(\bm{\theta} + \alpha\mathbf{z}) - \mathcal{L}(\bm{\theta} - \alpha\mathbf{z})}{2\alpha}\mathbf{z} \approx \mathbf{z}\mathbf{z}^\top \nabla\mathcal{L}(\bm{\theta}),
\end{equation}
where $\mathbf{z}$ is a multivariate variable with zero mean and unit variance, e.g., following the multivariate Gaussian distribution, and $\alpha$ is a perturbation scale. Alternatively, we can use the one-sided formulation for this directional gradient:
\begin{equation}
    \nabla^{ZO}\mathcal{L}(\bm{\theta}) = \frac{\mathcal{L}(\bm{\theta} + \alpha\mathbf{z}) - \mathcal{L}(\bm{\theta})}{\alpha}\mathbf{z}.
    \label{eq.one-sided_zo}
\end{equation}
The above formulations are two-point estimations, requiring two forward propagations to obtain an unbiased estimation of the gradient.
\begin{lemma}\label{lemma1}
When $\mathbf{z}$ has i.i.d. components with zero mean and unit variance, in the limit $\alpha\rightarrow 0$, $\nabla^{ZO}\mathcal{L}(\bm{\theta})$ is an unbiased estimator of $\nabla\mathcal{L}(\bm{\theta})$, i.e., $\mathbb{E}_{\mathbf{z}}\left[\nabla^{ZO}\mathcal{L}(\bm{\theta})\right] = \nabla\mathcal{L}(\bm{\theta})$.
\end{lemma}

Considering biological plausibility and efficiency, estimation with a single forward pass is more appealing. Actually, when considering expectation over $\mathbf{z}$, $\mathcal{L}(\bm{\theta})$ in Eq.~(\ref{eq.one-sided_zo}) can be omitted. Therefore, we can obtain a single-point zeroth-order (ZO$_{\text{sp}}$) unbiased estimator:
\begin{equation}
    \nabla^{ZO_{sp}}\mathcal{L}(\bm{\theta}) = \frac{\mathcal{L}(\bm{\theta} + \alpha\mathbf{z})}{\alpha}\mathbf{z}.
    \label{eq.one-point_zo}
\end{equation}
\begin{lemma}\label{lemma2}
When $\mathbf{z}$ has i.i.d. components with zero mean and unit variance, in the limit $\alpha\rightarrow 0$, $\nabla^{ZO_{sp}}\mathcal{L}(\bm{\theta})$ is an unbiased estimator of $\nabla\mathcal{L}(\bm{\theta})$.
\end{lemma}

\vspace{-1mm}
The above formulation only requires a noise injection in the forward propagation, and the gradients can be estimated with a top-down feedback signal, as shown in Fig.~\ref{fig:illustration}(d). This is also similar to REINFORCE~\citep{williams1992simple} and Evolution Strategies~\citep{salimans2017evolution} in reinforcement learning, and is considered to be biologically plausible~\citep{fiete2006gradient}. It is believed that the brain is likely to employ perturbation methods for some kinds of learning~\citep{lillicrap2020backpropagation}.

However, zeroth-order methods usually suffer from a large variance, since two-point methods only estimate gradients in a random direction and the one-point formulation has even larger variances. Therefore they hardly work for general neural network training. In the following, we propose our pseudo-zeroth-order method to solve the problem, also only based on one forward propagation with noise injection and top-down feedback signals.

\vspace{-1mm}
\section{Online Pseudo-Zeroth-Order Training}\label{sec:method}

In this section, we introduce the proposed online pseudo-zeroth-order method. We first introduce the pseudo-zeroth-order formulation for neural network training in Section~\ref{sec:pzo}. Then in Section~\ref{sec:mfc}, we introduce momentum feedback connections in OPZO for error propagation with zeroth-order estimation of the model. In Section~\ref{sec:forward_online_training}, we demonstrate the combination with online training and a similar form as the three-factor Hebbian learning. Finally, we introduce more additional details in Section~\ref{sec:additional_details}.

\subsection{Pseudo-Zeroth-Order Formulation}\label{sec:pzo}

Since zeroth-order methods suffer from large variances, a natural thought is to reduce the variance. However, ZO methods only rely on a scalar feedback signal to act on the random direction $\mathbf{z}$, making it hard to improve gradient estimation. To this end, we introduce a pseudo-zeroth-order formulation. As we build our work on online training, we first focus on the condition of a single SNN time step.

Specifically, we decouple the model function $f(\cdot; \bm{\theta})$ and the loss function $\mathcal{L}(\cdot)$. For each input $\mathbf{x}$, the model outputs $\mathbf{o}=f(\mathbf{x}; \bm{\theta})$, and then the loss is calculated as $\mathcal{L}(\mathbf{o}, \mathbf{y}_{\mathbf{x}})$, where $\mathbf{y}_{\mathbf{x}}$ is the label for the input. Different from ZO methods that only leverage the function value of $\mathcal{L}\circ f$, we assume that the gradient of $\mathcal{L}(\cdot)$ can be easily calculated, while keeping the zeroth-order formulation for $f(\cdot; \bm{\theta})$. This is consistent with real settings where gradients of the loss function have easy closed-form formulation, e.g., for mean-square-error (MSE) loss, $\nabla_{\mathbf{o}} \mathcal{L}(\mathbf{o}, \mathbf{y}_{\mathbf{x}}) = \mathbf{o} - \mathbf{y}_{\mathbf{x}}$, and for cross-entropy (CE) loss with the softmax function $\sigma$, $\nabla_{\mathbf{o}} \mathcal{L}(\mathbf{o}, \mathbf{y}_{\mathbf{x}}) = \sigma(\mathbf{o}) - \mathbf{y}_{\mathbf{x}}$, while gradients of $f(\cdot; \bm{\theta})$ are hard to compute due to biological plausibility issues or non-differentiability of spikes. 

With this formulation, we can consider feedback (error) signals $\mathbf{e} = \nabla_{\mathbf{o}} \mathcal{L}(\mathbf{o}, \mathbf{y}_{\mathbf{x}})$ that carries more information than a single value of $\mathcal{L}\circ f(\mathbf{x})$, potentially encouraging techniques for variance reduction. In the following, we introduce momentum feedback connections to directly propagate feedback signals to hidden layers for gradient estimation.

\subsection{Momentum Feedback Connections}\label{sec:mfc}

We motivate our method by first considering the directional gradient by the two-point estimation in Section~\ref{sec:zo}. With decoupled $f(\cdot; \bm{\theta})$ and $\mathcal{L}(\cdot)$ as in the pseudo-zeroth-order formulation and Taylor expansion of $\mathcal{L}(\cdot)$, Eq.~\ref{eq.one-sided_zo} turns into:
\vspace{-1.5mm}
\begin{equation}
    \nabla^{ZO}_{\bm{\theta}}\mathcal{L} \approx \frac{\left<\nabla_{\mathbf{o}}\mathcal{L}(\mathbf{o}, \mathbf{y}_{\mathbf{x}}), \tilde{\mathbf{o}} - \mathbf{o}\right>}{\alpha}\mathbf{z} = \mathbf{z} \frac{{\Delta \mathbf{o}}^\top}{\alpha} \nabla_{\mathbf{o}}\mathcal{L}(\mathbf{o}, \mathbf{y}_{\mathbf{x}}),
    \label{eq.two-point-pzo}
\end{equation}
where $\mathbf{o}=f(\mathbf{x}; \bm{\theta}), \tilde{\mathbf{o}}=f(\mathbf{x}; \bm{\theta}+\alpha\mathbf{z})$, and $\Delta \mathbf{o} = \tilde{\mathbf{o}} - \mathbf{o}$. This can be viewed as propagating the error signal with a connection weight $\mathbf{z} \frac{{\Delta \mathbf{o}}^\top}{\alpha}$. To reduce the variance introduced by the random direction $\mathbf{z}$, we introduce momentum feedback connections across different iterations and propagate errors as:
\vspace{-5mm}
\begin{equation}
\begin{aligned}
    \mathbf{M} &\coloneqq \lambda \mathbf{M} + (1 - \lambda) \mathbf{z} \frac{{\Delta \mathbf{o}}^\top}{\alpha}, \\
    \nabla^{PZO}_{\bm{\theta}}\mathcal{L} &= \mathbf{M} \nabla_{\mathbf{o}}\mathcal{L}(\mathbf{o}, \mathbf{y}_{\mathbf{x}}).
\end{aligned}
\end{equation}

\vspace{-3mm}
The momentum feedback connections can take advantage of different sampled directions $\mathbf{z}$, largely alleviating the variance caused by random directions. 

The above formulation only considers the directional gradient with two-point estimation, while we are more interested in methods with a single forward pass. Actually, $\mathbf{z} \frac{{\Delta \mathbf{o}}^\top}{\alpha}$ can be viewed as an unbiased estimator of $\mathbb{E}_{\mathbf{x}} \left[{\mathbf{J}_f^\top(\mathbf{x})}\right]$, where $\mathbf{J}_f(\mathbf{x})$ is the Jacobian of $f$ evaluated at $\mathbf{x}$, and $\mathbf{M}$ can be viewed as approximating it with moving average. Therefore, we can similarly use a one-point method and $\mathbf{M}$ has similar effects:
\vspace{-2mm}
\begin{equation}
    \mathbf{M} \coloneqq \lambda \mathbf{M} + (1 - \lambda) \mathbf{z} \frac{{\tilde{\mathbf{o}}}^\top}{\alpha},
\vspace{-1mm}
\end{equation}
where $\frac{{\tilde{\mathbf{o}}}^\top}{\alpha}$ is also an unbiased estimator of $\mathbb{E}_{\mathbf{x}} \left[{\mathbf{J}_f^\top(\mathbf{x})}\right]$.
\begin{lemma}\label{lemma3}
When $\mathbf{z}$ has i.i.d. components with zero mean and unit variance independent of inputs, in the limit $\alpha\rightarrow 0$, $\mathbf{z} \frac{{\Delta \mathbf{o}}^\top}{\alpha}$ and $\mathbf{z} \frac{{\tilde{\mathbf{o}}}^\top}{\alpha}$ are unbiased estimators of ${\mathbf{J}_f^\top(\mathbf{x})}$ given $\mathbf{x}$, and further, are unbiased estimators of $\mathbb{E}_{\mathbf{x}} \left[{\mathbf{J}_f^\top(\mathbf{x})}\right]$.
\end{lemma}

\vspace{-2mm}
This leads to our method as shown in Fig.~\ref{fig:illustration}(e). During forward propagation, a random noise $\alpha\mathbf{z}$ is injected for each layer, and momentum feedback connections are updated based on $\mathbf{z}$ and the model output $\tilde{\mathbf{o}}$ (information from pre- and post-synaptic neurons). Then errors are propagated through the connections to each layer\footnote{A small difference is that errors here are $\nabla_{\mathbf{o}}\mathcal{L}(\tilde{\mathbf{o}}, \mathbf{y}_{\mathbf{x}})$. 
This can be viewed as performing Taylor expansion at $\tilde{\mathbf{o}}$ for Eq.~\ref{eq.two-point-pzo}}. We consider node perturbation which is superior to weight perturbation\footnote{Node perturbation estimates gradients for $\mathbf{x}^{l+1}$ considering the neural network formulation $\mathbf{x}^{l+1}=\phi\left(\mathbf{W}^l\mathbf{x}^l\right)$ and calculates gradients as $\nabla_{\mathbf{W}^l}\mathcal{L}=\left(\nabla_{\mathbf{x}^{l+1}}\mathcal{L}\odot \phi'\left(\mathbf{W}^l\mathbf{x}^l\right)\right){\mathbf{x}^l}^\top$, which has a smaller variance than directly estimating gradients for weights.}~\citep{lillicrap2020backpropagation}, then it has a similar form as the popular DFA~\citep{nokland2016direct}, while our feedback weight is not a random matrix but the estimated Jacobian (Fig.~\ref{fig:illustration}(c,e)).

Then we analyze some properties of momentum feedback connections. We assume that $\mathbf{M}$ can quickly converge to the estimated $\mathbb{E}_{\mathbf{x}} \left[{\mathbf{J}_f^\top(\mathbf{x})}\right]$ up to small errors $\bm{\epsilon}$ compared with the optimization of parameters\footnote{Intuitively, for the given parameters of the model, $\mathbf{M}$ is just approximating a linear matrix with projection to different directions, which can converge more quickly than the highly non-convex optimization of neural networks. For the gradually evolving parameters, since the weight update for overparameterized neural networks is small for each iteration, we can also expect $\mathbf{M}$ to quickly fit the expectation of Jacobian.}, and we focus on gradient estimation with $\mathbf{M}=\mathbb{E}_{\mathbf{x}} \left[{\mathbf{J}_f^\top(\mathbf{x})}\right]+\mathbf{\epsilon}$. We show that it can largely reduce the variance of the single-point zeroth-order method (the proof and discussions are in Appendix~\ref{supsec:proof}).
\begin{proposition}\label{pro1}
Let $d$ denote the dimension of $\bm{\theta}$, $m$ denote the dimension of $\mathbf{o}$ ($m \ll d$), $B$ denote the mini-batch size, $\beta=\mathrm{Var}\left[z_i^2\right]$, 
$V_{\bm{\theta}}=\frac{1}{d}\sum_i\mathrm{Var}\left[(\nabla_{\bm{\theta}}\mathcal{L}_{\mathbf{x}})_i\right], S_{\bm{\theta}}=\frac{1}{d}\sum_i\mathbb{E}\left[(\nabla_{\bm{\theta}}\mathcal{L}_{\mathbf{x}})_i\right]^2, V_{L}=\mathrm{Var}\left[\mathcal{L}_{\mathbf{x}}\right], S_{L}=\mathbb{E}\left[\mathcal{L}_{\mathbf{x}}\right]^2$, $V_{\mathbf{o}}=\frac{1}{m}\sum_i\mathrm{Var}\left[(\nabla_{\mathbf{o}}\mathcal{L}_{\mathbf{x}})_i\right], S_{\mathbf{o}}=\frac{1}{m}\sum_i\mathbb{E}\left[(\nabla_{\mathbf{o}}\mathcal{L}_{\mathbf{x}})_i\right]^2$, where $\mathcal{L}_{\mathbf{x}}$ is the sample loss for input $\mathbf{x}$, and $\nabla_{\bm{\theta}}\mathcal{L}_{\mathbf{x}}$ and $\nabla_{\mathbf{o}}\mathcal{L}_{\mathbf{x}}$ are the sample gradient for $\bm{\theta}$ and $\mathbf{o}$, respectively.
We further assume that the small error $\bm{\epsilon}$ has i.i.d. components with zero mean and variance $V_{\bm{\epsilon}}$, and let $V_{\mathbf{o}, \mathbf{M}}=\frac{1}{d}\sum_{i,j}\mathrm{Var}\left[(\nabla_{\mathbf{o}}\mathcal{L}_{\mathbf{x}})_j\right](\mathbb{E}_{\mathbf{x}} \left[{\mathbf{J}_f^\top(\mathbf{x})}\right])_{i,j}$. 
Then the average variance of the single-point zeroth-order method is: 
$\frac{1}{B}\left((d+\beta)V_{\bm{\theta}}+(d+\beta-1)S_{\bm{\theta}}+\frac{1}{\alpha^2}V_L+\frac{1}{\alpha^2}S_L\right)+O(\alpha^2),$ 
while that of the pseudo-zeroth-order method is: 
$\frac{1}{B}\left(mV_{\bm{\epsilon}}V_{\mathbf{o}}+mV_{\bm{\epsilon}}S_{\mathbf{o}}+V_{\mathbf{o}, \mathbf{M}}\right).$
\end{proposition}

\begin{remark}
    $V_{\bm{\theta}}$ corresponds to the sample variance of spatial BP, and $V_{\mathbf{o}, \mathbf{M}}$ would be at a similar scale as $V_{\bm{\theta}}$ (see discussions in Appendix~\ref{supsec:proof}). Since $V_{\bm{\epsilon}}$ is expected to be very small, the results show that the single-point zeroth-order estimation has at least $d$ times larger variance than BP, while the pseudo-zeroth-order method can significantly reduce the variance, which is also verified in experiments.
\end{remark}

Besides the variance, another question is that momentum connections would take the expectation of the Jacobian over data $\mathbf{x}$, which can introduce bias into the gradient estimation. This is due to the data-dependent non-linearity that leads to a data-dependent Jacobian, which can be a shared problem for direct error feedback methods without layer-by-layer spatial BP. Despite the bias, we show that under certain conditions, the estimated gradient can still provide a descent direction (the proof and discussions are in Appendix~\ref{supsec:proof}).
\begin{proposition}\label{pro2}
    Suppose that ${\mathbf{J}_f^\top(\mathbf{x})}$ is $L_J$-Liptschitz continuous and $\mathbf{e}(\mathbf{x})$ is $L_e$-Liptschitz continuous, $\mathbf{x}_i$ is uniformly distributed, when $\left\lVert \mathbb{E}_{\mathbf{x}_i} \left[{\mathbf{J}_f^\top(\mathbf{x}_i)}\mathbf{e}(\mathbf{x}_i)\right] \right\rVert > \frac{1}{2}L_JL_e\Delta_{\mathbf{x}} + e_{\bm{\epsilon}}$, where $\Delta_{\mathbf{x}}=\mathbb{E}_{\mathbf{x}_i, \mathbf{x}_j}\left[\left\lVert \mathbf{x}_i - \mathbf{x}_j \right\rVert^2\right]$ and $e_{\bm{\epsilon}}=\left\lVert\bm{\epsilon}\mathbb{E}_{\mathbf{x}_i} \left[\mathbf{e}(\mathbf{x}_i)\right]\right\rVert$, we have $\left<\mathbb{E}_{\mathbf{x}_i} \left[{\mathbf{J}_f^\top(\mathbf{x}_i)}\mathbf{e}(\mathbf{x}_i)\right], \mathbb{E}_{\mathbf{x}_i} \left[\mathbf{M}\mathbf{e}(\mathbf{x}_i)\right] \right> > 0$.
\end{proposition}

There is a minor point that the above analyses mainly focus on differentiable functions while spiking neural networks are usually discrete and non-differentiable. Actually, under the stochastic setting, spiking neurons can be differentiable (see Appendix~\ref{supsec:sg} for details), and the deterministic setting may be roughly viewed as a special case. In practice, our method is not limited to differentiable functions but can estimate the expectation of the Jacobians of (potentially non-differentiable) black-box functions with zeroth-order methods. Our theoretical analyses mainly consider the well-defined continuous condition for insights. The pseudo-zeroth-order formulation may also be generalized to flexible hybrid zeroth-order and first-order systems for error signal propagation. In this work, we mainly focus on training SNNs, and we further introduce the combination with online training in the following.

\subsection{Online Pseudo-Zeroth-Order Training}\label{sec:forward_online_training}

We build the above pseudo-zeroth-order approach on online training methods to deal with spatial and temporal credit assignments. As introduced in Section~\ref{sec:snn}, we consider OTTT~\citep{xiao2022online} and replace its backpropagated instantaneous gradient with our estimated gradient based on direct top-down feedback. Then the update for synaptic weights has a similar form as the three-factor Hebbian learning~\citep{fremaux2016neuromodulated} based on a direct top-down global modulator:
\vspace{-0.5mm}
\begin{equation}
    \Delta W_{i,j} \propto \hat{a}_i[t] \psi(u_j[t]) \left(-g_j^t\right),
\end{equation}
where $W_{i,j}$ is the weight from neuron $i$ to $j$, $\hat{a}_i[t]$ is the presynaptic activity trace, $\psi(u_j[t])$ is a local surrogate derivative for the change rate of the postsynaptic activity~\citep{xiao2022online}, and $g_j^t$ is the global top-down error (gradient) modulator. Here we leverage the local surrogate derivative because it can be well-defined under the stochastic setting (see Appendix~\ref{supsec:sg} for details) and better fits the biological three-factor Hebbian rule.

For potentially asynchronous neuromorphic computing, there may be a delay in the propagation of error signals. \citet{xiao2022online} show that with convergent inputs and a certain surrogate derivative, it is still theoretically effective for the gradient under the delay $\Delta t$, i.e., the update is based on $\hat{a}_i[t+\Delta t] \psi(u_j[t+\Delta t]) g_j^t$. Alternatively, more eligibility traces can be used to store the local information, e.g., $\hat{a}_i[t] \psi(u_j[t])$, and induce weight updates when the top-down signal arrives~\citep{bellec2020solution}. Our method shares these properties and we do not model delays in experiments for the efficiency of simulations.

Moreover, the direct error propagations to different layers as well as the update of feedback connections in our method can be parallel, which can better take advantage of parallel neuromorphic computing than layer-by-layer spatial BP.

\subsection{Additional Details}\label{sec:additional_details}

\paragraph{Combination with Local Learning}
There can be both global and local signals for learning in biological systems, and local learning (LL) can improve global learning approximation methods~\citep{ren2023scaling}. Our proposed method can be combined with local learning for improvement, and we consider introducing local readout layers for local supervision. We will add a fully connected readout for each layer with supervised loss. Additionally, for deeper networks, we can also introduce intermediate global learning (IGL) that propagates global signals from a middle layer to previous ones with OPZO. More details can be found in Appendix~\ref{supsec:implementation_details}.  

\vspace{-2.5mm}
\paragraph{About Noise Injection} 
By default, we sample $\mathbf{z}$ from the Gaussian distribution. As sampling from the Gaussian distribution may pose computational requirements for hardware, we can also consider easier distributions. For example, the Rademacher distribution, which takes $1$ and $-1$ both with the probability $0.5$, also meets the requirements. Additionally, $\mathbf{z}$ is by default added to the neural activities for gradient estimation based on node perturbation. To further prevent noise perturbation from interfering with sparse spike-driven forward propagation of SNNs for energy efficiency, we may empirically change the noise perturbation after neuronal activities to perturbation before neurons (i.e., perturb on membrane potentials), while maintaining local surrogate derivatives for the spiking function. We will show in experiments that OPZO is robust to these noise injection settings.

\vspace{-2.5mm}
\paragraph{Antithetic Variables across Time Steps}
Compared with the two-point zeroth-order estimation, the considered one-point method can have a much larger variance. To further reduce the variance, we can leverage antithetic $\mathbf{z}$, i.e., $\mathbf{z}$ and $-\mathbf{z}$, for every two time steps of SNNs. Since SNNs naturally have multiple time steps and the inputs for different time steps usually belong to the same object with similar distributions, this approach may roughly approximate the two-point formulation without additional costs.

\begin{table*} [ht]
    \centering
    \small
    \caption{Accuracy (\%) of different spatial credit assignment methods with online training on various datasets.}
    \begin{tabular}{ccccccc}
        \toprule[1pt]
        Method & N-MNIST & DVS-Gesture & DVS-CIFAR10 & MNIST & CIFAR-10 & CIFAR-100 \\
        \midrule[0.5pt]
        Spatial BP \& SG & \underline{98.15$\pm$0.05} & \underline{95.72$\pm$0.33} & \underline{75.43$\pm$0.39} & \textbf{98.38$\pm$0.02} & \textbf{89.99$\pm$0.06} & \textbf{64.82$\pm$0.09} \\
        \midrule[0.5pt]
        DFA & 97.98$\pm$0.03 & 91.67$\pm$0.75 & 60.60$\pm$0.67 & 98.05$\pm$0.04 & 79.90$\pm$0.15 & 49.50$\pm$0.13 \\
        DFA (w/ LL) & / & 91.43$\pm$0.59 & 61.77$\pm$0.62 & / & 82.38$\pm$0.22 & 54.76$\pm$0.21 \\
        \midrule[0.5pt]
        ZO$_{\text{sp}}$ & 72.90$\pm$1.14 & 23.73$\pm$2.38 & 31.67$\pm$0.24 & 86.53$\pm$0.11 & 49.04$\pm$0.63 & 22.26$\pm$0.51 \\
        OPZO & \textbf{98.27$\pm$0.04} & 94.33$\pm$0.16 & 72.77$\pm$0.82 & \underline{98.34$\pm$0.10} & 85.74$\pm$0.15 & 60.93$\pm$0.16 \\
        OPZO (w/ LL) & / & \textbf{96.06$\pm$0.33} & \textbf{77.47$\pm$0.12} & / & \underline{89.42$\pm$0.16} & \underline{64.77$\pm$0.16} \\
        \bottomrule[1pt]
    \end{tabular}
    \label{table:performance}
\end{table*}

\section{Experiments}

In this section, we conduct experiments on both neuromorphic and static datasets with fully connected (FC) and convolutional (Conv) neural networks to demonstrate the effectiveness of the proposed OPZO method. For N-MNIST and MNIST, we leverage FC networks with two hidden layers composed of 800 neurons, and for DVS-CIFAR10, DVS-Gesture, CIFAR-10, and CIFAR-100, we leverage 5-layer convolutional networks. We will also consider a deeper 9-layer convolutional network, as well as fine-tuning ResNet-34 on ImageNet under noise. We take $T=30$ time steps for N-MNIST, $T=20$ for DVS-Gesture, $T=10$ for DVS-CIFAR10, and $T=6$ time steps for static datasets, following previous works~\citep{xiao2022online,zhang2020temporal}. More training details can be found in Appendix~\ref{supsec:implementation_details}.

\subsection{Comparison on Various Datasets}

We first compare the proposed OPZO with other spatial credit assignment methods on various datasets in Table~\ref{table:performance}, and all methods are based on the online training method OTTT~\citep{xiao2022online} under the same settings. For FC networks, since there are only two hidden layers, we do not consider local learning settings. As shown in the results, the ZO$_{\text{sp}}$ method fails to effectively optimize neural networks, while OPZO significantly improves the results, achieving performance at a similar level as spatial BP with SG. DFA with random feedback has a large gap with spatial BP, especially on convolutional networks, while OPZO can achieve much better results. When combined with local learning, OPZO has about the same performance as and even outperforms spatial BP with SG on neuromorphic datasets. These results demonstrate the effectiveness of OPZO for training SNNs to promising performance in a more biologically plausible and neuromorphic-friendly approach, paving paths for direct on-chip training of SNNs.

Note that our method is a different line from most recent works with state-of-the-art performance~\citep{kim2022exploring, li2023seenn,zhou2023spikformer} which are based on spatial and temporal BP with SG and focus on model improvement. We aim to develop alternatives to BP that adhere to neuromorphic properties, focusing on more biologically plausible and hardware-friendly training algorithms. So we mainly compare different spatial credit assignment methods under the same settings.

\begin{figure*}[ht]
    \centering
    \includegraphics[width=\textwidth]{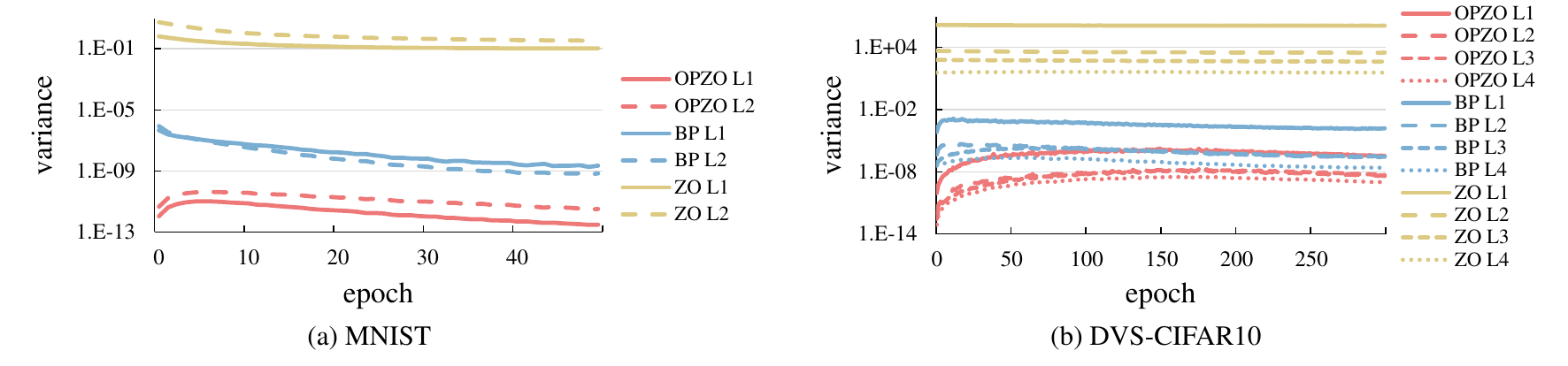}
    \vspace{-6mm}
    \caption{Results of gradient variances of OPZO, spatial BP with SG, and ZO$_{\text{sp}}$ on different datasets. ``L$i$'' denotes the $i$-th layer.}
    \label{fig:grad_var}
\vspace{-2mm}
\end{figure*}

\subsection{Gradient Variance}

We analyze the gradient variance of different methods to verify that our method can effectively reduce variance for effective training. As shown in Fig.~\ref{fig:grad_var}, the variance of ZO$_{\text{sp}}$ is several orders larger than spatial BP with SG, leading to the failure of effective training. OPZO can largely reduce the variance, which is consistent with our theoretical analysis. As in practice, we remove the factor $1/\alpha$ for the calculation of $\mathbf{M}$ (see Appendix~\ref{supsec:implementation_details} for details), which will scale gradients by the factor $\alpha$, the practical gradient variance of OPZO can be smaller than spatial BP with SG, but they are at a comparable level overall.

\begin{table} [h]
    \centering
    \small
    \tabcolsep=1mm
    \caption{Accuracy (\%) of OPZO on CIFAR-10 with different kinds of noise injection.}
    \begin{tabular}{ccc}
        \toprule[1pt]
        Noise distribution & Perturb after neuron & Perturb before neuron \\
        \midrule[0.5pt]
        Gaussian & 85.73$\pm$0.15 & 84.37$\pm$0.13 \\
        \midrule[0.5pt]
        Rademacher & 85.69$\pm$0.17 & 84.03$\pm$0.23 \\
        \bottomrule[1pt]
    \end{tabular}
    \label{table:noise}
\end{table}

\subsection{Effectiveness for Different Noise Injection}

Then we verify the effectiveness of OPZO for different noise injection settings as introduced in Section~\ref{sec:additional_details}. As shown in Table~\ref{table:noise}, the results under different noise distributions and injection positions are similar, demonstrating the robustness of OPZO for different settings.

\begin{table} [h]
    \centering
    \small
    \tabcolsep=4mm
    \caption{Accuracy (\%) of different methods with a deeper network.}
    \begin{tabular}{ccc}
        \toprule[1pt]
        Method & DVS-Gesture & CIFAR-100 \\
        \midrule[0.5pt]
        Spatial BP \& SG & 94.10$\pm$1.02 & 65.96$\pm$0.52 \\
        \midrule[0.5pt]
        DFA (w/ LL) & 93.40$\pm$0.49 & 52.94$\pm$0.20 \\
        DFA (w/ LL\&IGL) & 93.29$\pm$0.33 & 54.17$\pm$0.54 \\
        \midrule[0.5pt]
        OPZO (w/ LL) & 95.83$\pm$0.85 & 65.87$\pm$0.13 \\
        OPZO (w/ LL\&IGL) & \textbf{96.88$\pm$0.28} & \textbf{66.13$\pm$0.15} \\
        \bottomrule[1pt]
    \end{tabular}
    \label{table:deeper}
\end{table}

\begin{table} [h]
    \centering
    \small
    \caption{Accuracy (\%) of different methods for fine-tuning ResNet-34 on ImageNet under noise. BP refers to spatial BP with SG.}
    \begin{tabular}{cccccc}
        \toprule[1pt]
        Noise scale & Direct test & BP & DFA & ZO$_{\text{sp}}$ & OPZO \\
        \midrule[0.5pt]
        0.1 & 61.13 & 63.91 & 61.20 & 52.42 & 63.39 \\
        \midrule[0.5pt]
        0.15 & 54.01 & 62.13 & 54.59 & 30.32 & 60.96 \\
        \bottomrule[1pt]
    \end{tabular}
    \label{table:imagenet}
\end{table}

\subsection{Deeper Networks}

We further consider deeper networks and larger datasets. We first perform experiments on DVS-Gesture and CIFAR-100 with a deeper 9-layer convolutional network. As introduced in Section~\ref{sec:additional_details}, for deeper networks, we can introduce techniques including local learning and intermediate global learning. As shown in Table~\ref{table:deeper}, OPZO can also achieve similar performance as or outperform spatial BP with SG and significantly outperform DFA combined with these techniques.

We also conduct experiments for fine-tuning ResNet-34 on ImageNet under noise. This task is on the ground that there can be hardware mismatch, e.g., hardware noise, for deploying SNN models trained on common devices to neuromorphic hardware~\citep{yang2022training,cramer2022surrogate}, and we may expect direct on-chip fine-tuning to better deal with the problem. Our method is more plausible and efficient for on-chip learning than spatial BP with SG and may be combined with other works aiming at high-performance training on common devices in this scenario. We fine-tune a pre-trained NF-ResNet-34 model released by \citet{xiao2022online}, whose original test accuracy is $65.15\%$, under the noise injection setting with different scales. As shown in Table~\ref{table:imagenet}, OPZO can successfully fine-tune such large-scale models, while DFA and ZO$_{\text{sp}}$ fail. Spatial BP is less biologically plausible and is neuromorphic-unfriendly, so its results are only for reference. The results show that our method can scale to large-scale settings.

\subsection{Training Costs}

Finally, we analyze and compare the computational costs of different methods. We consider the estimation of the costs on potential neuromorphic hardware which is the target of neuromorphic computing with SNNs. Since biological systems leverage unidirectional local synapses, spatial BP (if we assume it to be possible for weight transport and separate forward-backward stage) should maintain additional backward connections between successive layers for layer-by-layer error backpropagation, leading to high memory and operation costs, as shown in Table~\ref{table:cost_neuromorphic}. Differently, DFA and OPZO maintain direct top-down feedback with much smaller costs, and they are also parallelizable for different layers. ZO$_{\text{sp}}$ may have even lower costs as only a scalar signal is propagated, which is shared by all neurons, but it cannot perform effective learning in practice. Also note that different from previous zeroth-order methods that require multiple forward propagations or forward gradient methods that require an additional heterogeneous signal propagation, our method only needs one common forward propagation with noise injection and direct top-down feedback, keeping lower operation costs which are about the same as DFA. We also provide training costs on common devices such as GPU in Appendix~\ref{supsec:additional_results}, and our method is comparable to spatial BP and DFA since GPU do not follow neuromorphic properties and we do not perform low-level code optimization as BP. It can be interesting future work to consider applications to neuromorphic hardware that is still under development~\citep{davies2021taking,schuman2022opportunities}.

For more experimental results, please refer to Appendix~\ref{supsec:additional_results}.

\begin{table} [t]
    \centering
    \small
    \tabcolsep=1mm
    \caption{Estimation of training costs on potential neuromorphic hardware for $N$-hidden-layer neural networks with $n$ neurons for hidden layers and $m$ neurons for the output layer, where $m \ll n$. The costs focus on the error backward procedure. ``*'' denotes parallelizable for different layers.}
    \begin{tabular}{ccc}
        \toprule[1pt]
        Method & Memory & Operations \\
        \midrule[0.5pt]
        BP (if possible) & $O\left((N-1)n^2+mn\right)$ & $O\left((N-1)n^2+mn\right)$ \\
        \midrule[0.5pt]
        DFA* & $O\left(Nmn\right)$ & $O\left(Nmn\right)$ \\
        \midrule[0.5pt]
        ZO$_{\text{sp}}$* & $O(Nn)$ & $O(Nn)$ \\
        \midrule[0.5pt]
        OPZO* & $O\left(Nmn\right)$ & $O\left(Nmn\right)$ \\
        \bottomrule[1pt]
    \end{tabular}
    \label{table:cost_neuromorphic}
\end{table}

\section{Conclusion}

In this work, we propose a new online pseudo-zeroth-order method for training spiking neural networks in a more biologically plausible and neuromorphic-hardware-friendly way, with low costs and comparable performance to spatial BP with SG. OPZO performs spatial credit assignment by a single common forward propagation with noise injection and direct top-down feedback based on momentum feedback connections, avoiding drawbacks of spatial BP, solving the large variance problem of zeroth-order methods, and significantly outperforming random feedback methods. Combining online training, OPZO has a similar form as three-factor Hebbian learning with direct top-down modulations, taking a step forward towards on-chip SNN training. Extensive experiments demonstrate the effectiveness and robustness of OPZO for fully connected and convolutional networks on static and neuromorphic datasets as well as larger models and datasets, and show the efficiency of OPZO with low estimated training costs.



\bibliography{arxiv}
\bibliographystyle{icml}

\newpage
\appendix
\onecolumn
\section{Detailed Proofs}\label{supsec:proof}

In this section, we provide proofs for lemmas and propositions in the main text.

\subsection{Proof of Lemma~\ref{lemma1} and Lemma~\ref{lemma2}}

\begin{proof}
In the limit $\alpha \rightarrow 0$, $\nabla^{ZO}\mathcal{L}(\bm{\theta}) =  \mathbf{z}\mathbf{z}^\top \nabla\mathcal{L}(\bm{\theta})$. Since $\mathbf{z}$ has i.i.d. components with zero mean and unit variance, we have $\mathbb{E}\left[\mathbf{z}\mathbf{z}^\top\right]=\mathbf{I}$. Therefore, $\mathbb{E}_{\mathbf{z}}\left[\nabla^{ZO}\mathcal{L}(\bm{\theta})\right]=\nabla\mathcal{L}(\bm{\theta})$.

Moreover, $\mathbb{E}_{\mathbf{z}}\left[\nabla^{ZO}\mathcal{L}(\bm{\theta})\right]=\mathbb{E}_{\mathbf{z}}\left[\frac{\mathcal{L}(\bm{\theta} + \alpha\mathbf{z}) - \mathcal{L}(\bm{\theta})}{\alpha}\mathbf{z}\right]=\mathbb{E}_{\mathbf{z}}\left[\frac{\mathcal{L}(\bm{\theta} + \alpha\mathbf{z})}{\alpha}\mathbf{z}\right] - \frac{\mathcal{L}(\mathbf{\theta})}{\alpha}\mathbb{E}_{\mathbf{z}}\left[\mathbf{z}\right] = \mathbb{E}_{\mathbf{z}}\left[\frac{\mathcal{L}(\bm{\theta} + \alpha\mathbf{z})}{\alpha}\mathbf{z}\right] = \mathbb{E}_{\mathbf{z}}\left[\nabla^{ZO_{sp}}\mathcal{L}(\bm{\theta})\right]$. Therefore, $\mathbb{E}_{\mathbf{z}}\left[\nabla^{ZO_{sp}}\mathcal{L}(\bm{\theta})\right] = \nabla\mathcal{L}(\bm{\theta})$.
\end{proof}

\subsection{Proof of Lemma~\ref{lemma3}}

\begin{proof}
In the limit $\alpha \rightarrow 0$, $\mathbf{z} \frac{{\Delta \mathbf{o}}^\top}{\alpha} = \mathbf{z} \left(\mathbf{J}_f(\mathbf{x})\mathbf{z}\right)^\top = \mathbf{z}\mathbf{z}^\top \mathbf{J}_f^\top(\mathbf{x})$. Since $\mathbb{E}\left[\mathbf{z}\mathbf{z}^\top\right]=\mathbf{I}$, we have $\mathbb{E}_{\mathbf{z}}\left[\mathbf{z} \frac{{\Delta \mathbf{o}}^\top}{\alpha} \middle| \mathbf{x}\right]=\mathbf{J}_f^\top(\mathbf{x})$. Then $\mathbb{E}_{\mathbf{x}, \mathbf{z}}\left[\mathbf{z} \frac{{\Delta \mathbf{o}}^\top}{\alpha}\right]=\mathbb{E}_{\mathbf{x}}\left[\mathbf{J}_f^\top(\mathbf{x})\right]$.

Also, $\mathbb{E}_{\mathbf{z}}\left[\mathbf{z} \frac{{\tilde{\mathbf{o}}}^\top}{\alpha} \middle| \mathbf{x}\right]=\mathbb{E}_{\mathbf{z}}\left[\mathbf{z} \frac{{\Delta \mathbf{o}}^\top}{\alpha} + \mathbf{z} \frac{{\mathbf{o}}^\top}{\alpha} \middle| \mathbf{x}\right]=\mathbb{E}_{\mathbf{z}}\left[\mathbf{z} \frac{{\Delta \mathbf{o}}^\top}{\alpha} \middle| \mathbf{x}\right] + \mathbb{E}_{\mathbf{z}}\left[\mathbf{z}\right]\frac{{\mathbf{o}}^\top}{\alpha} = \mathbb{E}_{\mathbf{z}}\left[\mathbf{z} \frac{{\Delta \mathbf{o}}^\top}{\alpha} \middle| \mathbf{x}\right]=\mathbf{J}_f^\top(\mathbf{x})$. Therefore, $\mathbb{E}_{\mathbf{x}, \mathbf{z}}\left[\mathbf{z} \frac{{\tilde{\mathbf{o}}}^\top}{\alpha}\right]=\mathbb{E}_{\mathbf{x}}\left[\mathbf{J}_f^\top(\mathbf{x})\right]$.
\end{proof}

\subsection{Proof of Proposition~\ref{pro1}}

\begin{proof}
We first consider the average variance of the two-point ZO estimation $\nabla_{\bm{\theta}}^{ZO}\mathcal{L}= \mathbf{z}\mathbf{z}^\top \nabla_{\bm{\theta}}\mathcal{L}+O(\alpha)$. Since $\mathrm{Var}(xy)=\mathrm{Var}(x)\mathrm{Var}(y)+\mathrm{Var}(x){\mathbb{E}(y)}^2+\mathrm{Var}(y){\mathbb{E}(x)}^2$ for independent $x$ and $y$, and $\mathbb{E}[z_i^2]=\mathrm{Var}[z_i]+{\mathbb{E}[z_i]}^2=1$, for each element of the gradient under sample $\mathbf{x}$, we have:
\begin{equation}
\begin{aligned}
    \mathrm{Var}\left[\left(\nabla_{\bm{\theta}}^{ZO}\mathcal{L}_{\mathbf{x}}\right)_i\right] &= \mathrm{Var}\left[\sum_{j=1}^d z_i z_j \left(\nabla_{\bm{\theta}}\mathcal{L}_{\mathbf{x}}\right)_j\right] + O(\alpha^2) \\
    &=\mathrm{Var}\left[z_i^2 \left(\nabla_{\bm{\theta}}\mathcal{L}_{\mathbf{x}}\right)_i\right] + \sum_{j\neq i} \mathrm{Var}\left[z_i z_j \left(\nabla_{\bm{\theta}}\mathcal{L}_{\mathbf{x}}\right)_j\right] + O(\alpha^2) \\
    &=\mathrm{Var}\left[z_i^2\right] \mathrm{Var}\left[\left(\nabla_{\bm{\theta}}\mathcal{L}_{\mathbf{x}}\right)_i\right] + \mathrm{Var}\left[z_i^2\right] {\mathbb{E}\left[\left(\nabla_{\bm{\theta}}\mathcal{L}_{\mathbf{x}}\right)_i\right]}^2 + \mathrm{Var}\left[\left(\nabla_{\bm{\theta}}\mathcal{L}_{\mathbf{x}}\right)_i\right] {\mathbb{E}\left[z_i^2\right]}^2\\
    &\quad +\sum_{j\neq i}\left( \mathrm{Var}\left[z_iz_j\right] \mathrm{Var}\left[\left(\nabla_{\bm{\theta}}\mathcal{L}_{\mathbf{x}}\right)_j\right] + \mathrm{Var}\left[z_iz_j\right] {\mathbb{E}\left[\left(\nabla_{\bm{\theta}}\mathcal{L}_{\mathbf{x}}\right)_j\right]}^2 + \mathrm{Var}\left[\left(\nabla_{\bm{\theta}}\mathcal{L}_{\mathbf{x}}\right)_j\right] {\mathbb{E}\left[z_iz_j\right]}^2 \right) + O(\alpha^2)\\
    &=(\beta+1) \mathrm{Var}\left[\left(\nabla_{\bm{\theta}}\mathcal{L}_{\mathbf{x}}\right)_i\right] + \beta{\mathbb{E}\left[\left(\nabla_{\bm{\theta}}\mathcal{L}_{\mathbf{x}}\right)_i\right]}^2 + \sum_{j\neq i}\left( \mathrm{Var}\left[\left(\nabla_{\bm{\theta}}\mathcal{L}_{\mathbf{x}}\right)_j\right] + {\mathbb{E}\left[\left(\nabla_{\bm{\theta}}\mathcal{L}_{\mathbf{x}}\right)_j\right]}^2 \right) + O(\alpha^2)\\
    &=\beta\mathrm{Var}\left[\left(\nabla_{\bm{\theta}}\mathcal{L}_{\mathbf{x}}\right)_i\right] + (\beta-1){\mathbb{E}\left[\left(\nabla_{\bm{\theta}}\mathcal{L}_{\mathbf{x}}\right)_i\right]}^2 + \sum_{j=1}^d\left( \mathrm{Var}\left[\left(\nabla_{\bm{\theta}}\mathcal{L}_{\mathbf{x}}\right)_j\right] + {\mathbb{E}\left[\left(\nabla_{\bm{\theta}}\mathcal{L}_{\mathbf{x}}\right)_j\right]}^2 \right) + O(\alpha^2).
\end{aligned}
\end{equation}

Taking the average of all elements, we obtain the average variance for each sample (denoted as $\mathrm{mVar}$):

\begin{equation}
\begin{aligned}
    \mathrm{mVar}\left[\nabla_{\bm{\theta}}^{ZO}\mathcal{L}_{\mathbf{x}}\right] &= \frac{1}{d} \sum_{i=1}^d \mathrm{Var}\left[\left(\nabla_{\bm{\theta}}^{ZO}\mathcal{L}_{\mathbf{x}}\right)_i\right]\\
    &= \frac{\beta}{d} \sum_{i=1}^d \mathrm{Var}\left[\left(\nabla_{\bm{\theta}}\mathcal{L}_{\mathbf{x}}\right)_i\right] + \frac{\beta-1}{d} \sum_{i=1}^d {\mathbb{E}\left[\left(\nabla_{\bm{\theta}}\mathcal{L}_{\mathbf{x}}\right)_i\right]}^2 + \sum_{j=1}^d\left( \mathrm{Var}\left[\left(\nabla_{\bm{\theta}}\mathcal{L}_{\mathbf{x}}\right)_j\right] + {\mathbb{E}\left[\left(\nabla_{\bm{\theta}}\mathcal{L}_{\mathbf{x}}\right)_j\right]}^2 \right) + O(\alpha^2)\\
    &= (d+\beta)V_{\bm{\theta}} + (d+\beta-1)S_{\bm{\theta}} + O(\alpha^2).
\end{aligned}
\end{equation}

For gradient calculation with batch size $B$, the sample variance can be reduced by $B$ times, resulting in the average variance $\frac{1}{B}\left((d+\beta)V_{\bm{\theta}} + (d+\beta-1)S_{\bm{\theta}}\right) + O(\alpha^2)$.

Then we can derive the average variance of the single-point ZO estimation $\nabla_{\bm{\theta}}^{ZO_{sp}}\mathcal{L}=\nabla_{\bm{\theta}}^{ZO}\mathcal{L}+\frac{\mathcal{L}_{\mathbf{x}}}{\alpha}\mathbf{z}$ for each sample:

\begin{equation}
\begin{aligned}
    \mathrm{mVar}\left[\nabla_{\bm{\theta}}^{ZO_{sp}}\mathcal{L}_{\mathbf{x}}\right] &= \mathrm{mVar}\left[\nabla_{\bm{\theta}}^{ZO}\mathcal{L}_{\mathbf{x}}\right] + \mathrm{mVar}\left[\frac{\mathcal{L}_{\mathbf{x}}}{\alpha}\mathbf{z}\right] \\
    &= (d+\beta)V_{\bm{\theta}} + (d+\beta-1)S_{\bm{\theta}} + O(\alpha^2) + \frac{1}{\alpha^2} \left(\mathrm{Var}\left[\mathcal{L}_{\mathbf{x}}\right]\mathrm{Var}\left[z_i\right] + \mathrm{Var}\left[\mathcal{L}_{\mathbf{x}}\right]{\mathbb{E}\left[z_i\right]}^2 + \mathrm{Var}\left[z_i\right]{\mathbb{E}\left[\mathcal{L}_{\mathbf{x}}\right]}^2\right) \\
    &= (d+\beta)V_{\bm{\theta}} + (d+\beta-1)S_{\bm{\theta}} + \frac{1}{\alpha^2}V_L + \frac{1}{\alpha^2}S_L + O(\alpha^2).
\end{aligned}
\end{equation}

For batch size $B$, the average variance is $\frac{1}{B}\left((d+\beta)V_{\bm{\theta}}+(d+\beta-1)S_{\bm{\theta}}+\frac{1}{\alpha^2}V_L+\frac{1}{\alpha^2}S_L\right)+O(\alpha^2)$.

Next, we turn to the average variance of the pseudo-zeroth-order method $\nabla^{PZO}_{\bm{\theta}}\mathcal{L} = \mathbf{M} \nabla_{\mathbf{o}}\mathcal{L}_{\mathbf{x}} = \left(\mathbb{E}_{\mathbf{x}} \left[{\mathbf{J}_f^\top(\mathbf{x})}\right])+\bm{\epsilon}\right)\nabla_{\mathbf{o}}\mathcal{L}_{\mathbf{x}}$. For each element, we have:

\begin{equation}
\begin{aligned}
    \mathrm{Var}\left[\left(\nabla_{\bm{\theta}}^{PZO}\mathcal{L}_{\mathbf{x}}\right)_i\right] &= \mathrm{Var}\left[\sum_{j=1}^m \left(\mathbb{E}_{\mathbf{x}} \left[{\mathbf{J}_f^\top(\mathbf{x})}\right])\right)_{i,j}\left(\nabla_{\mathbf{o}}\mathcal{L}_{\mathbf{x}}\right)_j\right] + \mathrm{Var}\left[\sum_{j=1}^m \epsilon_{i,j}\left(\nabla_{\mathbf{o}}\mathcal{L}_{\mathbf{x}}\right)_j\right] \\
    &= \sum_{j=1}^m \left(\mathbb{E}_{\mathbf{x}} \left[{\mathbf{J}_f^\top(\mathbf{x})}\right])\right)_{i,j} \mathrm{Var}\left[\left(\nabla_{\mathbf{o}}\mathcal{L}_{\mathbf{x}}\right)_j\right] + \sum_{j=1}^m \left(V_{\bm{\epsilon}}\mathrm{Var}\left[\left(\nabla_{\mathbf{o}}\mathcal{L}_{\mathbf{x}}\right)_j\right] + V_{\bm{\epsilon}}{\mathbb{E}\left[\left(\nabla_{\mathbf{o}}\mathcal{L}_{\mathbf{x}}\right)_j\right]}^2\right).
\end{aligned}
\end{equation}

Taking the average of all elements, we have the average variance for each sample:

\begin{equation}
\begin{aligned}
    \mathrm{mVar}\left[\nabla_{\bm{\theta}}^{PZO}\mathcal{L}_{\mathbf{x}}\right] &= \frac{1}{d}\sum_{i=1}^d\sum_{j=1}^m \left(\mathbb{E}_{\mathbf{x}} \left[{\mathbf{J}_f^\top(\mathbf{x})}\right])\right)_{i,j} \mathrm{Var}\left[\left(\nabla_{\mathbf{o}}\mathcal{L}_{\mathbf{x}}\right)_j\right] + \sum_{j=1}^m \left(V_{\bm{\epsilon}}\mathrm{Var}\left[\left(\nabla_{\mathbf{o}}\mathcal{L}_{\mathbf{x}}\right)_j\right] + V_{\bm{\epsilon}}{\mathbb{E}\left[\left(\nabla_{\mathbf{o}}\mathcal{L}_{\mathbf{x}}\right)_j\right]}^2\right) \\
    &= mV_{\bm{\epsilon}}V_{\mathbf{o}}+mV_{\bm{\epsilon}}S_{\mathbf{o}}+V_{\mathbf{o}, \mathbf{M}}.
\end{aligned}
\end{equation}

Then for batch size $B$, the average variance is $\frac{1}{B}\left(mV_{\bm{\epsilon}}V_{\mathbf{o}}+mV_{\bm{\epsilon}}S_{\mathbf{o}}+V_{\mathbf{o}, \mathbf{M}}\right)$.
\end{proof}

\begin{remark}
    $\beta=\mathrm{Var}\left[z_i^2\right]=\mathbb{E}(z_i^4)-{\mathbb{E}(z_i^2)}^2=\mathbb{E}(z_i^4)-1$ depends on the distribution of $z_i$. For the Gaussian distribution, $\mathbb{E}(z_i^4)=3$ and therefore $\beta=2$. For the Rademacher distribution, $\mathbb{E}(z_i^4)=1$ and therefore $\beta=0$.
\end{remark}

\begin{remark}
    The zero mean assumption on the small error $\bm{\epsilon}$ is reasonable since $\mathbf{z}\frac{{\tilde{\mathbf{o}}}^\top}{\alpha}$ is an unbiased estimator for $\mathbb{E}_{\mathbf{x}}\left[\mathbf{J}_f^\top(\mathbf{x})\right]$, so the expectation of the error can be expected to be zero.
\end{remark}

\begin{remark}
    $V_{\bm{\theta}}$ and $V_{\mathbf{o}, \mathbf{M}}$ may not be directly compared considering the complex network function, but we may make a brief analysis under some simplifications. For $\left(\nabla_{\bm{\theta}}\mathcal{L}_{\mathbf{x}}\right)_i=\left(\mathbf{J}_f^\top(\mathbf{x})\nabla_{\mathbf{o}}\mathcal{L}_{\mathbf{x}}\right)_i$, let $\mathbf{J}_{i,j}$ and $\nabla_j$ denote $\left(\mathbf{J}_f^\top(\mathbf{x})\right)_{i,j}$ and $\left(\nabla_{\mathbf{o}}\mathcal{L}_{\mathbf{x}}\right)_j$ for short, we have $\mathrm{Var}\left[\left(\nabla_{\bm{\theta}}\mathcal{L}_{\mathbf{x}}\right)_i\right] = \mathrm{Var}\left[\sum_{j=1}^m \mathbf{J}_{i,j}\nabla_j\right] = \sum_j \mathrm{Var}\left[\mathbf{J}_{i,j}\nabla_j\right] + \sum_{j_1,j_2}\mathrm{Cov}\left[\mathbf{J}_{i,j_1}\nabla_{j_1}, \mathbf{J}_{i,j_2}\nabla_{j_2}\right] = \sum_j\left[\mathrm{Var}\left[\nabla_j\right]\mathbb{E}\left[\mathbf{J}_{i,j}^2\right] + \mathrm{Var}\left[\mathbf{J}_{i,j}\right]\mathbb{E}\left[\nabla_j\right]^2 + 2\mathrm{Cov}\left[\mathbf{J}_{i,j}, \nabla_j\right]\mathbb{E}\left[\mathbf{J}_{i,j}\right]\mathbb{E}\left[\nabla_j\right]\right] + \sum_{j_1,j_2}\mathrm{Cov}\left[\mathbf{J}_{i,j_1}\nabla_{j_1}, \mathbf{J}_{i,j_2}\nabla_{j_2}\right]$. If we ignore covariance terms and assume $\mathbb{E}\left[\nabla_j\right]=0$, this is simplified to $\sum_j\mathrm{Var}\left[\nabla_j\right]\mathbb{E}\left[\mathbf{J}_{i,j}^2\right]$, and then $V_{\bm{\theta}}$ is approximated as $\frac{1}{d}\sum_{i,j}\mathrm{Var}\left[\nabla_j\right]\mathbb{E}\left[\mathbf{J}_{i,j}^2\right]$, which has a similar form as $V_{\mathbf{o}, \mathbf{M}}=\frac{1}{d}\sum_{i,j}\mathrm{Var}\left[(\nabla_{\mathbf{o}}\mathcal{L}_{\mathbf{x}})_j\right](\mathbb{E}_{\mathbf{x}} \left[{\mathbf{J}_f^\top(\mathbf{x})}\right])_{i,j}$ except that the second moment is considered. Under this condition, the scales of $V_{\mathbf{o}, \mathbf{M}}$ and $V_{\bm{\theta}}$ may slightly differ considering the scale of elements of $\mathbf{J}_f^\top(\mathbf{x})$, but overall, $V_{\mathbf{o}, \mathbf{M}}$ would be at a similar scale as $V_{\bm{\theta}}$ compared with the variances of the zeroth-order methods that are at least $d$ times larger which is proportional to the number of intermediate neurons.
\end{remark}

\subsection{Proof of Proposition~\ref{pro2}}

\begin{proof}
Since ${\mathbf{J}_f^\top(\mathbf{x})}$ is $L_J$-Liptschitz continuous and $\mathbf{e}(\mathbf{x})$ is $L_e$-Liptschitz continuous, we have $\left\lVert\mathbf{J}_f^\top(\mathbf{x}_i) - \mathbf{J}_f^\top(\mathbf{x}_j)\right\rVert \leq L_J \left\lVert \mathbf{x}_i - \mathbf{x}_j \right\rVert$, $\left\lVert\mathbf{e}(\mathbf{x}_i) - \mathbf{e}(\mathbf{x}_j)\right\rVert \leq L_e \left\lVert \mathbf{x}_i - \mathbf{x}_j \right\rVert$. Then with the equation that $\frac{1}{2n^2}\sum_{i,j}(a_i-a_j)(b_i-b_j) = \frac{1}{n}\sum_i a_ib_i - \frac{1}{n^2}\sum_{i,j} a_ib_j$, we have 
\begin{equation}
\begin{aligned}
    &\left\lVert \mathbb{E}_{\mathbf{x}_i} \left[{\mathbf{J}_f^\top(\mathbf{x}_i)}\mathbf{e}(\mathbf{x}_i)\right] - \mathbb{E}_{\mathbf{x}_i} \left[\left(\mathbb{E}_{\mathbf{x}_j}\left[{\mathbf{J}_f^\top(\mathbf{x}_j)}\right]+\bm{\epsilon}\right)\mathbf{e}(\mathbf{x}_i)\right] \right\rVert\\
    =\quad &\left\lVert \frac{1}{n}\sum_{\mathbf{x}_i} \widetilde{\mathbf{J_f}}(\mathbf{x}_i)\mathbf{e}(\mathbf{x}_i) - \left(\frac{1}{n}\sum_{\mathbf{x}_i}\widetilde{\mathbf{J_f}}(\mathbf{x}_i)\right)\left(\frac{1}{n}\sum_{\mathbf{x}_i}\mathbf{e}(\mathbf{x}_i)\right) - \bm{\epsilon}\mathbb{E}_{\mathbf{x}_i} \left[\mathbf{e}(\mathbf{x}_i)\right] \right\rVert\\
    =\quad & \left\lVert \frac{1}{2n^2} \sum_{\mathbf{x}_i, \mathbf{x}_j} \left(\widetilde{\mathbf{J_f}}(\mathbf{x}_i) - \widetilde{\mathbf{J_f}}(\mathbf{x}_j)\right) \left(\mathbf{e}(\mathbf{x}_i) - \mathbf{e}(\mathbf{x}_j)\right) - \bm{\epsilon}\mathbb{E}_{\mathbf{x}_i} \left[\mathbf{e}(\mathbf{x}_i)\right] \right\rVert\\
    \leq\quad & \frac{1}{2n^2} \sum_{\mathbf{x}_i, \mathbf{x}_j} \left\lVert \left(\widetilde{\mathbf{J_f}}(\mathbf{x}_i) - \widetilde{\mathbf{J_f}}(\mathbf{x}_j)\right) \right\rVert \left\lVert \left(\mathbf{e}(\mathbf{x}_i) - \mathbf{e}(\mathbf{x}_j)\right) \right\rVert + \left\lVert  \bm{\epsilon}\mathbb{E}_{\mathbf{x}_i} \left[\mathbf{e}(\mathbf{x}_i)\right] \right\rVert\\
    \leq\quad & \frac{1}{2n^2} \sum_{\mathbf{x}_i, \mathbf{x}_j} L_JL_e {\left\lVert \mathbf{x}_i - \mathbf{x}_j \right\rVert}^2 + \left\lVert  \bm{\epsilon}\mathbb{E}_{\mathbf{x}_i} \left[\mathbf{e}(\mathbf{x}_i)\right] \right\rVert\\
    =\quad &\frac{1}{2} L_J L_e \Delta_{\mathbf{x}} + e_{\bm{\epsilon}}\\
    <\quad & \left\lVert \mathbb{E}_{\mathbf{x}_i} \left[{\mathbf{J}_f^\top(\mathbf{x}_i)}\mathbf{e}(\mathbf{x}_i)\right] \right\rVert.
\end{aligned}
\end{equation}
Therefore, 
\begin{equation}
\begin{aligned}
    &\left<\mathbb{E}_{\mathbf{x}_i} \left[{\mathbf{J}_f^\top(\mathbf{x}_i)}\mathbf{e}(\mathbf{x}_i)\right], \mathbb{E}_{\mathbf{x}_i} \left[\mathbf{M}\mathbf{e}(\mathbf{x}_i)\right] \right>\\ 
    =\quad &{\left\lVert \mathbb{E}_{\mathbf{x}_i} \left[{\mathbf{J}_f^\top(\mathbf{x}_i)}\mathbf{e}(\mathbf{x}_i)\right] \right\rVert}^2 - \left<\mathbb{E}_{\mathbf{x}_i} \left[{\mathbf{J}_f^\top(\mathbf{x}_i)}\mathbf{e}(\mathbf{x}_i)\right], \mathbb{E}_{\mathbf{x}_i} \left[{\mathbf{J}_f^\top(\mathbf{x}_i)}\mathbf{e}(\mathbf{x}_i)\right] - \mathbb{E}_{\mathbf{x}_i} \left[\mathbf{M}\mathbf{e}(\mathbf{x}_i)\right]\right>\\
    \geq\quad &{\left\lVert \mathbb{E}_{\mathbf{x}_i} \left[{\mathbf{J}_f^\top(\mathbf{x}_i)}\mathbf{e}(\mathbf{x}_i)\right] \right\rVert}^2 - \left\lVert \mathbb{E}_{\mathbf{x}_i} \left[{\mathbf{J}_f^\top(\mathbf{x}_i)}\mathbf{e}(\mathbf{x}_i)\right] \right\rVert \left\lVert \mathbb{E}_{\mathbf{x}_i} \left[{\mathbf{J}_f^\top(\mathbf{x}_i)}\mathbf{e}(\mathbf{x}_i)\right] - \mathbb{E}_{\mathbf{x}_i} \left[\left(\mathbb{E}_{\mathbf{x}_j}\left[{\mathbf{J}_f^\top(\mathbf{x}_j)}\right]+\bm{\epsilon}\right)\mathbf{e}(\mathbf{x}_i)\right] \right\rVert\\
    >\quad &0.
\end{aligned}
\end{equation}
\end{proof}

\begin{remark}
$L_J$ will depend on the non-linearity of the network, for example, $L_J=0$ for linear networks. This will influence the condition of effective descent direction considering the gradient scale as in the proposition. Note that these assumptions are not necessary premises, and we have verified the effectiveness of the method in experiments.
\end{remark}

\section{Introduction to Local Surrogate Derivatives under the Stochastic Spiking Setting}\label{supsec:sg}

In this section, we provide more introduction to the stochastic spiking setting, under which spiking neurons can be \textit{locally} differentiable and there exist \textit{local} surrogate derivatives.

Biological spiking neurons can be stochastic, where a neuron generates spikes following a Bernoulli distribution with the probability as the c.d.f. of a distribution w.r.t $u[t]-V_{th}$, indicating a higher probability for a spike with larger $u[t]-V_{th}$. That is, $s_i[t]$ is a random variable following a $\{0, 1\}$ valued Bernoulli distribution with the probability of $1$ as $p(s_i[t]=1)=F(u_i[t] - V_{th})$. With reparameterization, this can be formulated as $s_i[t] = H(u_i\left[t\right] - V_{th} - z_i)$ with a random noise variable $z_i$ that follows the distribution specified by $F$. Different $F$ corresponds to different distributions and noises. For example, the sigmoid function corresponds to a logistic noise, while the erf function corresponds to a Gaussian noise. 
Under the stochastic setting, the local surrogate derivatives can be introduced for the spiking function~\citep{shekhovtsov2021reintroducing,ma2023exploiting}.

Specifically, consider the objective function which should turn to the expectation over random variables under the stochastic model. Considering a one-hidden-layer network with one time step, with the input $\mathbf{x}$ connecting to $n$ spiking neurons by the weight $\mathbf{W}$ and the neurons connecting to an output readout layer by the weight $\mathbf{O}$. Different from deterministic models with the objective function $\mathbb{E}_{\mathbf{x}} [\mathcal{L}(\mathbf{s})]$, where $\mathbf{s}=H(\mathbf{u} - V_{th}), \mathbf{u}=\mathbf{W}\mathbf{x}$, under the stochastic setting, the objective is to minimize
\begin{equation}
    \mathbb{E}_{\mathbf{x}} [\mathbb{E}_{\mathbf{s}\sim p(\mathbf{s} | \mathbf{x}, \mathbf{W})} [\mathcal{L}(\mathbf{s})]].
\end{equation}
For this objective, the model can be differentiable and gradients can be derived~\citep{shekhovtsov2021reintroducing,ma2023exploiting}. We focus on the gradients of $\mathbf{u}$, which can be expressed as:
\begin{equation}
\begin{aligned}
    \frac{\partial}{\partial \mathbf{u}} \mathbb{E}_{\mathbf{s}\sim p(\mathbf{s} | \mathbf{W})} [\mathcal{L}(\mathbf{s})] &= \frac{\partial}{\partial \mathbf{u}} \sum_{\mathbf{s}} \left(\prod_i p(\mathbf{s}_i | \mathbf{W})\right) \mathcal{L}(\mathbf{s}) \\
    &= \sum_{\mathbf{s}}\sum_i \left( \prod_{i'\neq i}p(\mathbf{s}_{i'} | \mathbf{W}) \right) \left( \frac{\partial}{\partial \mathbf{u}} p(\mathbf{s}_i | \mathbf{W}) \right) \mathcal{L}(\mathbf{s}).
\end{aligned}
\label{eq:grad1}
\end{equation}
Then consider derandomization to perform summation over $s_i$ while keeping other random variables fixed~\citep{shekhovtsov2021reintroducing}. Let $\mathbf{s}_{\lnot i}$ denote other variables except $s_i$. Since $s_i$ is $\{0,1\}$ valued, given $\mathbf{s}_{\lnot i}$, we have
\begin{equation}
\begin{aligned}
    \sum_{s_i\in \{0, 1\}} \frac{\partial p(s_i | \mathbf{W})}{\partial \mathbf{u}} \mathcal{L}([\mathbf{s}_{\lnot i}, s_i]) &= \frac{\partial p(\mathbf{s}_i | \mathbf{W})}{\partial \mathbf{u}} \mathcal{L}(\mathbf{s}) + \frac{\partial (1 - p(\mathbf{s}_i | \mathbf{W}))}{\partial \mathbf{u}} \mathcal{L}(\mathbf{s}_{\downarrow i})\\
    &= \frac{\partial p(\mathbf{s}_i | \mathbf{W})}{\partial \mathbf{u}} \left(\mathcal{L}(\mathbf{s}) - \mathcal{L}(\mathbf{s}_{\downarrow i})\right), 
\end{aligned}
\end{equation}
where $\mathbf{s}$ is a random sample considering $s_i$ (the RHS is invariant of $s_i$), and $\mathbf{s}_{\downarrow i}$ denotes taking $\mathbf{s}_i$ as the other state for $\mathbf{s}$. Given that $\sum_{s_i} p(s_i | \mathbf{W})=1$, Eq.~(\ref{eq:grad1}) is equivalent to 
\begin{equation}
\begin{aligned}
    \frac{\partial}{\partial \mathbf{u}} \mathbb{E}_{\mathbf{s}\sim p(\mathbf{s} | \mathbf{W})} [\mathcal{L}(\mathbf{s})] &= \sum_i \sum_{\mathbf{s}_{\lnot i}} \left( \prod_{i'\neq i}p(\mathbf{s}_{i'} | \mathbf{W}) \right) \sum_{s_i} \left( \frac{\partial}{\partial \mathbf{u}} p(s_i | \mathbf{W}) \right) \mathcal{L}([\mathbf{s}_{\lnot i}, s_i])\\
    &= \sum_i \sum_{\mathbf{s}_{\lnot i}} \left( \prod_{i'\neq i}p(\mathbf{s}_{i'} | \mathbf{W}) \right) \sum_{s_i} p(s_i | \mathbf{W}) \frac{\partial p(\mathbf{s}_i | \mathbf{W})}{\partial \mathbf{u}} \left(\mathcal{L}(\mathbf{s}) - \mathcal{L}(\mathbf{s}_{\downarrow i})\right)\\
    &= \sum_{\mathbf{s}} \left( \prod_{i}p(\mathbf{s}_{i} | \mathbf{W}) \right) \sum_i \frac{\partial p(\mathbf{s}_i | \mathbf{W})}{\partial \mathbf{u}} \left(\mathcal{L}(\mathbf{s}) - \mathcal{L}(\mathbf{s}_{\downarrow i})\right)\\
    &= \mathbb{E}_{\mathbf{s}\sim p(\mathbf{s} | \mathbf{W})} \sum_i \frac{\partial p(\mathbf{s}_i | \mathbf{W})}{\partial \mathbf{u}} \left(\mathcal{L}(\mathbf{s}) - \mathcal{L}(\mathbf{s}_{\downarrow i})\right).
\end{aligned}
\label{eq:grad2}
\end{equation}
Taking one sample of $\mathbf{s}$ in each forward procedure allows the unbiased gradient estimation as the Monte Carlo method. In this equation, considering the probability distribution, we have: 
\begin{equation}
    \frac{\partial p(\mathbf{s}_i | \mathbf{W})}{\partial \mathbf{u}} = F'(\mathbf{u}, V_{th}), 
\end{equation}
where $F'$ is the derivative of $F$, corresponding to a \textit{local} surrogate gradient, e.g., the derivative of the sigmoid function, triangular function, etc. 

The term $\mathcal{L}(\mathbf{s}) - \mathcal{L}(\mathbf{s}_{\downarrow i})$ corresponds to the error, and the above derivation is also similar to REINFORCE~\citep{williams1992simple}. However, since it relies on derandomization, simultaneous perturbation is infeasible in this formulation, and for efficient simultaneous calculation of all components, we may follow previous works~\citep{shekhovtsov2021reintroducing} to tackle it by linear approximation: $\mathcal{L}(\mathbf{s}) - \mathcal{L}(\mathbf{s}_{\downarrow i})\approx \frac{\partial \mathcal{L}(\mathbf{s})}{\partial \mathbf{s}_i}$, enabling simultaneous calculation given a gradient $\frac{\partial \mathcal{L}(\mathbf{s})}{\partial \mathbf{s}}$. This approximation may introduce bias, while it can be small for over-parameterized neural networks with weights at the scale of $\frac{1}{\sqrt{d_n}}$, where $d_n$ is the neuron number. This means that for the elements of the readout $\mathbf{o}=\mathbf{O}\mathbf{s}$, flipping the state of $\mathbf{s}_i$ only has $O(\frac{1}{\sqrt{d_n}})$ influence.

The deterministic model may be viewed as a special case, e.g., with noise always as zero, and \citet{shekhovtsov2021reintroducing} show that the gradients under the deterministic setting can provide a similar ascent direction under certain conditions.

Therefore, spiking neurons can be differentiable under the stochastic setting and \textit{local} surrogate derivatives can be well-defined, supporting our formulation as introduced in the main text. Our pseudo-zeroth-order method approximates $\frac{\partial \mathcal{L}(\mathbf{s})}{\partial \mathbf{s}}$, fitting the above formulation. Also note that the above derivation of surrogate derivatives is \textit{local} for one hidden layer -- for multi-layer networks, while we may iteratively perform the above analysis to obtain the commonly used global surrogate gradients, there can be expanding errors through layer-by-layer propagation due to the linear approximation error. Differently, our OPZO performs direct error feedback, which may reduce such errors.

\section{More Implementation Details}\label{supsec:implementation_details}

\subsection{Local Learning}

For experiments with local learning, we consider local supervision with a fully connected readout for each layer. Specifically, for the output $\mathbf{s}^l$ of each layer, we calculate the local loss based on the readout $\mathbf{r}^l = \mathbf{R}^l\mathbf{s}^l$ as $\mathcal{L}(\mathbf{r}^l, \mathbf{y})$. Then the gradient for $\mathbf{s}^l$ is calculated by the local loss and added to the global gradient based on our OPZO method, which will update synaptic weights directly connected to the neurons. For simplicity, we assume the weight symmetry for propagating errors through $\mathbf{R}^l$ in code implementations, because for the single linear layer, weights can be easily learned to be symmetric~\cite{akrout2019deep}, e.g., based on a sleep phase with weight mirroring~\cite{akrout2019deep} or based on our zeroth-order formulation with noise injection. \citet{kaiser2020synaptic} also show that a fixed random matrix can be effective for such kind of local learning.

We also consider intermediate global learning (IGL) as a kind of local learning. That is, we choose a middle layer to perform readout for loss calculation, just as the last layer, and its direct feedback signal will be propagated to previous layers. For the experiments with a 9-layer network, we choose the middle layer as the fourth convolutional layer.

\subsection{Training Settings}

\subsubsection{Datasets}

We conduct experiments on N-MNIST~\citep{orchard2015converting}, DVS-Gesture~\citep{amir2017low}, DVS-CIFAR10~\citep{li2017cifar10}, MNIST~\citep{lecun1998gradient}, CIFAR-10 and CIFAR-100~\citep{krizhevsky2009learning}, as well as ImageNet~\cite{deng2009imagenet}.

\paragraph{N-MNIST} N-MNIST is a neuromorphic dataset converted from MNIST by a Dynamic Version Sensor (DVS), with the same number of training and testing samples as MNIST. Each sample consists of spike trains triggered by the intensity change of pixels when DVS scans a static MNIST image. There are two channels corresponding to ON- and OFF-event spikes, and the pixel dimension is expanded to $34\times34$ due to the relative shift of images. Therefore, the size of the spike trains for each sample is $34\times34\times2\times T$, where $T$ is the temporal length. The original data record $300ms$ with the resolution of $1\mu s$. We follow \citet{zhang2020temporal} to reduce the time resolution by accumulating the spike train within every $3ms$ and use the first 30 time steps. The license of N-MNIST is the Creative Commons Attribution-ShareAlike 4.0 license.

\paragraph{DVS-Gesture}
DVS-Gesture is a neuromorphic dataset recording 11 classes of hand gestures by a DVS camera. It consists of 1,176 training samples and 288 testing samples. Following \citet{Fang_2021_ICCV}, we pre-possess the data to integrate event data into 20 frames, and we reduce the spatial resolution to $48\times 48$ by interpolation. 
The license of DVS-Gesture is the Creative Commons Attribution 4.0 license.

\paragraph{DVS-CIFAR10}
DVS-CIFAR10 is the neuromorphic dataset converted from CIFAR-10 by DVS, which is composed of 10,000 samples, one-sixth of the original CIFAR-10. It consists of spike trains with two channels corresponding to ON- and OFF-event spikes. We split the dataset into 9000 training samples and 1000 testing samples as the common practice, and we reduce the temporal resolution by accumulating the spike events~\citep{Fang_2021_ICCV} into 10 time steps as well as the spatial resolution into $48\times48$ by interpolation. We apply the random cropping augmentation similar to CIFAR-10 to the input data and normalize the inputs based on the global mean and standard deviation of all time steps. The license of DVS-CIFAR10 is CC BY 4.0.

\paragraph{MNIST} MNIST consists of 10-class handwritten digits with 60,000 training samples and 10,000 testing samples. Each sample is a $28\times28$ grayscale image. We normalize the inputs based on the global mean and standard deviation, and convert the pixel value into a real-valued input current at every time step. The license of MNIST is the MIT License.

\paragraph{CIFAR-10} CIFAR-10 consists of 10-class color images of objects with 50,000 training samples and 10,000 testing samples. Each sample is a $32\times32\times3$ color image. We normalize the inputs based on the global mean and standard deviation, and apply random cropping, horizontal flipping, and cutout~\citep{devries2017improved} for data augmentation. The inputs to the first layer of SNNs at each time step are directly the pixel values, which can be viewed as a real-valued input current.

\paragraph{CIFAR-100} CIFAR-100 is a dataset similar to CIFAR-10 except that there are 100 classes of objects. It also consists of 50,000 training samples and 10,000 testing samples. We use the same pre-processing as CIFAR-10.

The license of CIFAR-10 and CIFAR-100 is the MIT License. 

\paragraph{ImageNet} ImageNet-1K is a dataset of color images with 1,000 classes of objects, containing 1,281,167 training samples and 50,000 validation images. We adopt the common pre-possessing strategies to first randomly resize and crop the input image to $224\times224$, and then normalize it after the random horizontal flipping data augmentation, while the testing images are first resized to $256\times256$ and center-cropped to $224\times224$, and then normalized. The inputs are also converted to a real-valued input current at each time step. The license of ImageNet is Custom (non-commercial).

\subsubsection{Training Details and Hyperparameters}

For SNN models, following the common practice, we leverage the accumulated membrane potential of the neurons at the last classification layer (which will not spike or reset) for classification, i.e., the classification during inference is based on the accumulated $\mathbf{u}^N[T]=\sum_{t=1}^T\mathbf{o}[t]$, where $\mathbf{o}[t]=\mathbf{W}^{N-1}\mathbf{s}^{N-1}[t]+\mathbf{b}^N$ which can be viewed as an output at each time step. The loss during training is calculated for each time step as $\mathcal{L}(\mathbf{o}[t], \mathbf{y})$ following the instantaneous loss in online training with the loss function as a combination of cross-entropy (CE) loss and mean-square-error (MSE) loss~\citep{xiao2022online}. For spiking neurons, $V_{th}=1$ and $\lambda=0.5$. We leverage the sigmoid-like local surrogate derivative, i.e., $\psi(u)=\frac{1}{a_1}\frac{e^{(V_{th}-u)/a_1}}{(1+e^{(V_{th}-u)/a_1})^2}$ with $a_1=0.25$. For convolutional networks, we apply the scaled weight standardization~\citep{brock2021characterizing} as in \citet{xiao2022online}.

For our OPZO method, as well as the ZO method in experiments, $\alpha$ is set as $0.2$ initially and linearly decays to $0.01$ through the epochs, in order to reduce the influence of stochasticness for forward propagation. For fine-tuning on ImageNet under noise, $\alpha$ is set as the noise scale, and we do not apply antithetic variables across time steps, in order to better fit the noisy test setting (perturbation noise is before the neuron). In practice, we remove the factor $1/\alpha$ for the calculation of $\mathbf{M}$, because in the single-point setting, the scale of $\tilde{\mathbf{o}}$ is larger than and not proportional to $\alpha$ (for the discrete spiking model, the scale of $\Delta \mathbf{o}$ can also be large and not proportional to $\alpha$). This only influences the estimated gradient with a scale $\alpha$, and may be offset by the adaptive optimizer. Additionally, viewing $\mathbf{M}$ as approximating its objective with gradient descent, the decreasing $\alpha$ may be viewed as the learning rate with a linear scheduler.

For N-MNIST and MNIST, we consider FC networks with two hidden layers composed of 800 neurons, and for DVS-CIFAR10, DVS-Gesture, CIFAR-10, and CIFAR-100, we consider 5-layer Conv networks (128C3-AP2-256C3-AP2-512C3-AP2-512C3-FC), or 9-layer Conv networks under the deeper network setting (64C3-128C3-AP2-256C3-256C3-AP2-512C3-512C3-AP2-512C3-512C3-FC). We train our models on common datasets by the AdamW optimizer with learning rate 2e-4 and weight decay 2e-4 (except for ZO, the learning rate is set as 2e-5 on DVS-CIFAR10, MNIST, CIFAR-10, and CIFAR-100 for better results). The batch size is set as 128 for most datasets and 16 for DVS-Gesture, and the learning rate is cosine annealing to 0.
For N-MNIST and MNIST, we train models by 50 epochs and we apply dropout with the rate 0.2 (except for ZO). 
For DVS-Gesture, DVS-CIFAR10, CIFAR-10, and CIFAR-100, we train models by 300 epochs. For DVS-CIFAR10, we apply dropout with the rate 0.1 (except for ZO). We set the momentum coefficient for momentum feedback connections as $\lambda=0.99999$ (except for DVS-Gesture, it is set as $\lambda=0.999999$ due to a smaller batch size), and for the combination with local learning, the local loss is scaled by $0.01$.

For fine-tuning ImageNet, the learning rate is set as 2e-6 (and 2e-7 for ZO) without weight decay, and the batch size is set as 64. The perturbation noise is before the neuron, i.e., added to the results after convolutional operations. For BP, we train 1 epoch. For DFA, ZO, and OPZO, we train 5 epochs. We observe that DFA and ZO fail after 1 epoch, so we only report the results after 1 epoch, and for OPZO, the results can continually improve, so we report the results after 5 epochs. The 1-epoch and 5-epoch results for OPZO are 63.04 and 63.39 under the noise scale of 0.1, and 59.50 and 60.96 under the noise scale of 0.15.

The code implementation is based on the PyTorch framework, and experiments are carried out on one NVIDIA GeForce RTX 3090 GPU. Experiments are based on 3 runs of experiments with the same random seeds 2022, 0, and 1.

For gradient variance experiments, the variances are calculated by the batch gradients in one epoch, i.e., $var = \frac{\sum\lVert \mathbf{g}_i - \overline{\mathbf{g}}\rVert^2}{n}$, where $\mathbf{g}_i$ is the batch gradient, $\overline{\mathbf{g}}$ is the average of batch gradients, and $n$ is the number of batches multiplied by the number of elements in the gradient vector. 

\section{Additional Results}\label{supsec:additional_results}

\subsection{Training Costs on GPU}

We provide a brief comparison of memory and time costs of different methods on GPU in Table~\ref{suptab:cost_gpu}. Our proposed OPZO has about the same costs as spatial BP with SG and DFA. If we exclude some code-level optimization and implement all methods in a similar fashion, DFA and OPZO are faster than spatial BP, which is consistent with the theoretical analysis of operation numbers. Note that this is only a brief comparison as we do not perform low-level code optimization for OPZO and DFA, for example, the direct feedback of OPZO and DFA to different layers can be parallel, and local learning for different layers can also theoretically be parallel, to further reduce the time. As described in the main text, the target of neuromorphic computing with SNNs would be potential neuromorphic hardware, and OPZO and DFA can have lower costs, while GPU generally does not follow the properties. Since neuromorphic hardware is still under development, we mainly simulate the experiments on GPU, and it can be future work to consider combination with hardware implementation.

\begin{table} [ht]
    \centering
    \caption{Brief comparison of training costs on GPU for CIFAR-10 with convolutional networks. $^\dag$ means manual implementation of spatial BP \& SG with layer-by-layer backpropagation, which is in a similar fashion as other methods. $^\ddag$ means using automatic differentiation implemented by PyTorch with low-level code optimizations.}
    \begin{tabular}{ccc}
        \toprule[1pt]
        Method & Memory & Time per epoch \\
        \midrule[0.5pt]
        Spatial BP \& SG & 2.8G$^\dag$ / 2.9G$^\ddag$ & 49s$^\dag$ / 45s$^\ddag$ \\
        DFA & 2.8G & 44s \\
        ZO$_{\text{sp}}$ & 2.8G & 46s \\
        OPZO & 2.9G & 46s \\
        \midrule[0.5pt]
        DFA (w/ LL) & 3.0G & 50s \\
        OPZO (w/ LL) & 3.1G & 51s \\
        \bottomrule[1pt]
    \end{tabular}
    \label{suptab:cost_gpu}
\end{table}

\subsection{Firing Rate and Synaptic Operations}

\begin{table} [ht]
    \centering
    \caption{The firing rate (fr) and synaptic operations (SynOp) induced by spikes for models trained by different methods on DVS-CIFAR10 and CIFAR-10.}
    \begin{tabular}{ccccccc}
        \multicolumn{7}{c}{\textbf{DVS-CIFAR10}} \\
        \toprule[1pt]
        Method & Layer1 fr & Layer2 fr & Layer3 fr & Layer4 fr & Total fr & SynOp \\
        \midrule[0.5pt]
        Spatial BP \& SG & 0.1763 & 0.1733 & 0.2394 & 0.3575 & 0.1904 & $1.42\times 10^9$ \\
        \midrule[0.5pt]
        DFA & 0.2693 & 0.4564 & 0.4783 & 0.4930 & 0.3574 & $2.91\times 10^9$ \\
        \midrule[0.5pt]
        DFA (w/ LL) & 0.2433 & 0.4531 & 0.4848 & 0.4919 & 0.3430 & $2.84\times 10^9$ \\
        \midrule[0.5pt]
        OPZO & 0.2435 & 0.3446 & 0.4212 & 0.4222 & 0.3021 & $2.41\times 10^9$ \\
        \midrule[0.5pt]
        OPZO (w/ LL) & 0.0406 & 0.0614 & 0.1451 & 0.2838 & \textbf{0.0691} & $\mathbf{0.53\times 10^9}$ \\
        \bottomrule[1pt]
    \end{tabular}
    \begin{tabular}{ccccccc}
        \multicolumn{7}{c}{\textbf{CIFAR-10}} \\
        \toprule[1pt]
        Method & Layer1 fr & Layer2 fr & Layer3 fr & Layer4 fr & Total fr & SynOp \\
        \midrule[0.5pt]
        Spatial BP \& SG & 0.2005 & 0.1679 & 0.1067 & 0.0493 & 0.1734 & $0.76\times 10^9$ \\
        \midrule[0.5pt]
        DFA & 0.1769 & 0.3787 & 0.4314 & 0.4149 & 0.2759 & $1.40\times 10^9$ \\
        \midrule[0.5pt]
        DFA (w/ LL) & 0.1196 & 0.3180 & 0.4089 & 0.3878 & 0.2235 & $1.16\times 10^9$ \\
        \midrule[0.5pt]
        OPZO & 0.1563 & 0.2861 & 0.3496 & 0.2754 & 0.2229 & $1.12\times 10^9$ \\
        \midrule[0.5pt]
        OPZO (w/ LL) & 0.0400 & 0.0670 & 0.1159 & 0.2157 & \textbf{0.0640} & $\mathbf{0.30\times 10^9}$ \\
        \bottomrule[1pt]
    \end{tabular}
    \label{suptab:firing_rate}
\end{table}

For event-driven SNNs, the energy costs on neuromorphic hardware are proportional to the spike count, or more precisely, synaptic operations induced by spikes. Therefore, we also compare the firing rate (i.e., average spike count per neuron per time step) and synaptic operations of the models trained by different methods. As shown in Table~\ref{suptab:firing_rate}, on both DVS-CIFAR10 and CIFAR-10, OPZO (w/ LL) achieves the lowest average total firing rate and synaptic operations, indicating the most energy efficiency. The results also demonstrate different spike patterns for models trained by different methods, and show that LL can significantly improve OPZO while can hardly improve DFA. It may indicate OPZO as a better more biologically plausible global learning method to be combined with local learning.

\subsection{Training Dynamics}

\begin{figure*}[ht]
    \centering
    \includegraphics[width=\textwidth]{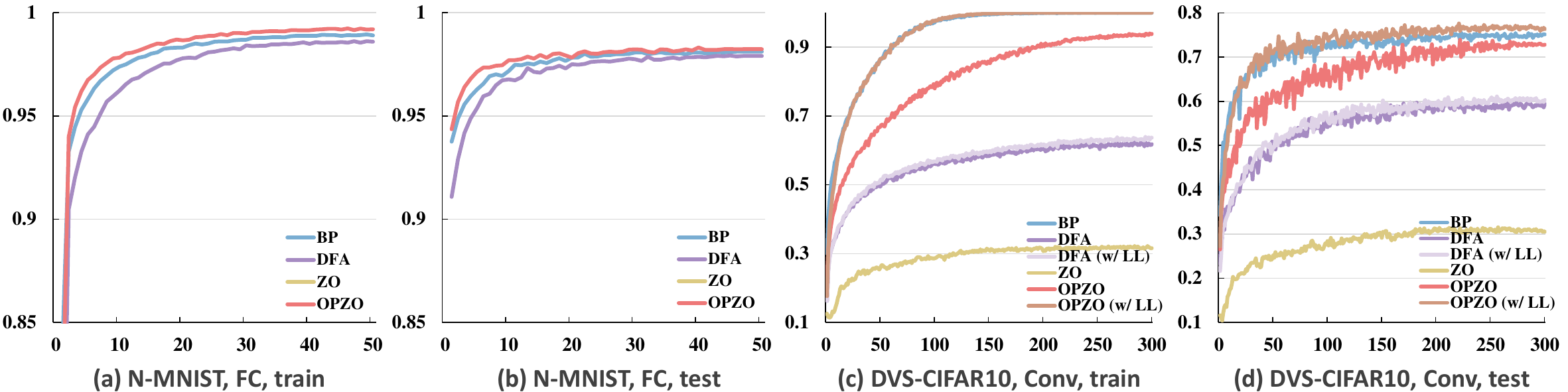}
    \caption{Training dynamics of different methods on N-MNIST and DVS-CIFAR10.}
    \label{fig:training_dynamics}
\end{figure*}

We present the training dynamics of different methods in Fig.~\ref{fig:training_dynamics}. For fully connected networks on N-MNIST, OPZO achieves a similar convergence speed as spatial BP with SG, which is better than DFA. For convolutional networks on DVS-CIFAR10, OPZO itself is slower than spatial BP with SG but performs much better than DFA, and when combined with local learning, OPZO (w/ LL) achieves a similar training convergence speed as BP as well as a better testing performance.

\subsection{More Comparisons}

\begin{table} [h]
    \centering
    \caption{Comparison results with DKP which updates error feedback matrices. Experiments are under the same random seed.}
    \begin{tabular}{ccccccc}
        \toprule[1pt]
        Method & N-MNIST & DVS-Gesture & DVS-CIFAR10 & MNIST & CIFAR-10 & CIFAR-100 \\
        \midrule[0.5pt]
        DFA & 97.93 & 91.32 & 59.70 & 98.00 & 79.84 & 49.68 \\
        \midrule[0.5pt]
        DKP & 97.81 & 51.39 & 39.40 & 98.14 & 81.12 & 53.35 \\
        \midrule[0.5pt]
        OPZO & 98.31 & 94.44 & 73.90 & 98.44 & 85.63 & 60.78 \\
	\bottomrule[1pt]
    \end{tabular}
\label{suptable:comparison_dkp}
\end{table}

We note that some previous works propose to update error feedback matrices for DFA, e.g., DKP~\citep{webster2020learning} that may have similar training costs as OPZO. DKP is based on the formulation of DFA and updates feedback weights similar to Kolen-Pollack learning, which calculates gradients for feedback weights by the product of the middle layer’s activation and the error from the top layer. Its basic thought is trying to keep the update direction of feedback and feedforward weights the same, but it may lack sufficient theoretical groundings. 

In this subsection, we compare DKP with DFA and the proposed OPZO method. As DKP is designed for ANN, we implement it for SNN with the adaptation of activations to pre-synaptic traces for feedback weight learning (similar to the update of feedforward weight). As shown in Table~\ref{suptable:comparison_dkp}, compared with DFA, DKP can have around 2-3\% performance improvement on CIFAR-10 and CIFAR-100, which is similar to the improvement in its paper. However, we observe that DKP cannot work well for neuromorphic datasets. And OPZO significantly outperforms both DKP and DFA on all datasets, also with more theoretical guarantees.


\end{document}